\documentclass[acmsmall]{acmart}

\AtBeginDocument{%
  \providecommand\BibTeX{{%
    \normalfont B\kern-0.5em{\scshape i\kern-0.25em b}\kern-0.8em\TeX}}}

\setcopyright{acmcopyright}
\copyrightyear{2022}
\acmYear{2022}
\acmDOI{}

\acmJournal{CSUR}
\acmVolume{}
\acmNumber{}
\acmArticle{}
\acmMonth{}

\usepackage{lscape}
\usepackage{longtable}
\usepackage{multirow}
\usepackage{subcaption}
\usepackage{hyperref}
\usepackage{xfrac}

\setcopyright{none}
\renewcommand\footnotetextcopyrightpermission[1]{} % removes footnote with conference information in first column
\begin{document}

%%
%% The ''title'' command has an optional parameter,
%% allowing the author to define a ''short title'' to be used in page headers.
\title{Socially Enhanced Situation Awareness from Microblogs using Artificial Intelligence: A Survey}

%%
%% The ''author'' command and its associated commands are used to define
%% ''authornote'' and ''authornotemark'' commands
\author{Rabindra Lamsal}
\email{rlamsal@student.unimelb.edu.au}
\orcid{0000-0002-2182-3001}
\author{Aaron Harwood}
\email{aharwood@unimelb.edu.au}
\author{Maria Rodriguez Read}
\email{maria.read@unimelb.edu.au}
\affiliation{%
  \institution{School of Computing and Information Systems, University of Melbourne}
  \streetaddress{Melbourne Connect, 700 Swanston Street}
  \city{Melbourne}
  \state{Victoria}
  \country{Australia}
  \postcode{3010}
  \thanks{This study was supported by the Melbourne Research Scholarship from the University of Melbourne, Australia.}
}

\renewcommand{\shortauthors}{Lamsal et al.}

\begin{abstract}
The rise of social media platforms provides an unbounded, infinitely rich source of aggregate knowledge of the world around us, both historic and real-time, from a human perspective. The greatest challenge we face is how to process and understand this raw and unstructured data, go beyond individual observations and see the ``big picture''---the domain of Situation Awareness. We provide an extensive survey of Artificial Intelligence research, focusing on microblog social media data with applications to Situation Awareness, that gives the seminal work and state-of-the-art approaches across six thematic areas: \emph{Crime}, \emph{Disasters}, \emph{Finance}, \emph{Physical Environment}, \emph{Politics}, and \emph{Health and Population}. We provide a novel, unified methodological perspective, identify key results and challenges, and present ongoing research directions.
\end{abstract}




\maketitle

\section{Introduction}
Being able to understand the state of the world around us---referred to as \emph{situation awareness}---is a fundamental capability for making effective decisions affecting our future. The role of ``Social Media'', sometimes attributed to as ``Web 2.0'', is at the forefront of this endeavour today. Social media refers to those internet and mobile-based services that facilitate online exchanges of conversations, social networking and forming online communities for contributing user-created content among their users~\cite{dewing2010social}. Arguably, the vast uptake of social media platforms has revolutionized the way people interact socially with each other. These platforms allow individuals, organizations and governments to interact with a large number of people, all in real-time, through the exchange of texts, photos, videos and social network cues such as liking and following.

The billions of active social media users per month~\cite{statista_2021} provide a big data processing challenge that is infeasible without employing high performance, large-scale distributed computational methods---with a rapidly increasing demand for \textit{machine learning}~\cite{bishop2006pattern} approaches. Machine Learning (ML) is a subset of \textit{Artificial Intelligence} (AI) that primarily focuses on making a system learn and improve itself without involving explicit programming, by utilizing concepts primarily from statistics, probability, linear algebra, and differential calculus. ML shares its application scope with many inter-disciplinary areas, including medical~\cite{kononenko2001machine}, linguistics~\cite{manning2015computational}, time series forecasting~\cite{ahmed2010empirical}, and computer vision~\cite{voulodimos2018deep}. The various learning paradigms associated with ML are discussed later in Section~\ref{background}.

\subsection{Situation awareness from social media}
Speed, transparency, and ubiquity, aided by the proliferation of mobile technology, are the main reasons for the growth of social media. For example, situations that would have otherwise remained relatively unknown for an indefinite period are now being reported and shared worldwide within minutes~\cite{mayfield2008social}. It literally takes only a few seconds to report a situation globally on social media: take a photo, write a few words and share it online. This reporting method is more efficient than individual cellular communication (e.g. making individual calls to friends and family) since social media effectively provides a public broadcast platform for individuals. Effective inclusion of information drawn from social media has become a necessity for government authorities to better understand the environment and people that they govern. The online interaction between users with the exchange of status updates, news stories, and other media has potential to be significantly advantageous for \textit{situation awareness}. 

As defined in~\cite{endsley2000theoretical}, situation awareness is a three-stage process: \textit{Perception}, \textit{Comprehension}, and \textit{Projection}.
In the \textit{Perception stage}, information regarding an event is collected and unwanted information is filtered out. The \textit{Comprehension stage} involves integrating multiple pieces of information collected from the earlier stage and studying its relevance and validity. The last stage of situation awareness is the \textit{Projection stage}, and it has been found that even experienced individuals rely on future projections heavily while dealing with situations. Projecting the current scenarios and dynamics to their future implications gives space for timely and effective decision-making.

A situation is traditionally perceived using observations, questionnaires, interviews, checklists and measurements~\cite{endsley2000theoretical}. Social media conversations in the form of text, photos, and videos overlap the observation, questionnaire and interview methods of collecting data. Previous studies have shown that such socially generated data contributes to a better understanding of an ongoing situation~\cite{imran2015processing}. For example, during a disastrous event people tend to use social media excessively, as they share their safety status and exchange conversations to query the safety status of their friends and family. People also share what they have seen, felt, or heard from others. During such critical hours of a disaster, the use of social media can peak to unprecedented levels, and based on these public conversations first responders and decision-makers can visualize a more comprehensive real-time picture of the situation to aid in formulating actionable plans.

\subsection{The social media spectrum}

\begin{figure}[t]
  \centering
  \includegraphics[width=\linewidth]{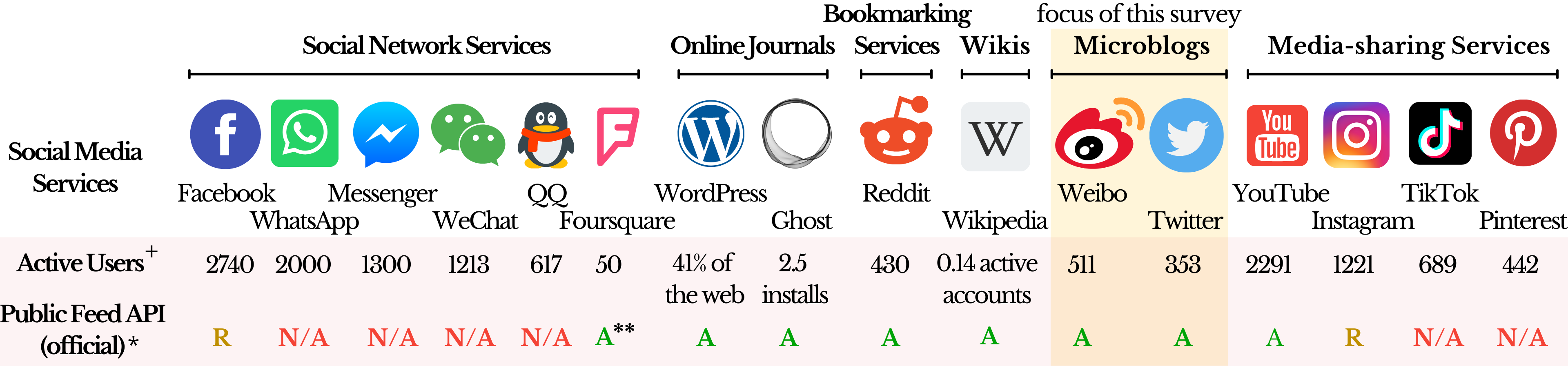}
  \caption{Classification of some popular~\cite{statista_2021} social media services. \textit{Notes:} $^{+}$in millions, unless mentioned otherwise (as of January 2021); $^{*}$R represents restricted usage (restricted to a limited set of publishers);  N/A represents not available;  A represents available; $^{**}$Places API for accessing global point of interest (POI) data.}
  \label{social-media-category}
\end{figure}

Figure~\ref{social-media-category} classifies some of the popular~\cite{statista_2021} social media services. \textit{Social Network Services} are platforms that allow users to create profiles and develop a network of connections so that their exchanged conversations are shared effectively within the network. Instant Messaging Services are a subclass of Social Network Services, considering their main feature of creating users' networks. \textit{Online Journals}, commonly referred to as ``blogs'', let their users publish stories/ideas usually in a reverse chronological order. \textit{Social Bookmarking Services} allow users to bookmark links to websites. These services can generate enormous traffic to the bookmarked web content. \textit{Microblogs} are online broadcasting services that allow users to post updates with a limited number of words or characters. \textit{Wikis} are collaborative platforms powered by wiki engines where their users can edit and manage the content. Unlike content management system such as WordPress, the content production in a wiki is not limited to a sole author. \textit{Media-Sharing Services} allow their users to share media contents such as images and videos. This classification does not have strict boundaries because one way or another, these services borrow features from the other side of the spectrum to keep their users engaged. Consider Twitter---it is a microblogging platform, but it also acts as a Social Network and also allows sharing of images and short videos just like Media-Sharing Services.

\subsubsection{Suitability for situation awareness}

To be an effective source of social media for enhancing situation awareness we consider three key attributes (shown in Figure~\ref{social-media-category}): the content should carry real-time descriptions, there should be a significantly large and active userbase, and the service must provide a public streaming API. With these criteria in mind, we focused this survey on Microblog services, primarily research that uses Twitter data. Some social media services may not have real-time attributes, i.e. their content is updated over a long time intervals, but they still contribute to enhancing situation awareness by enriching the data obtained from their real-time peers. Consider, for example, an event associated with the \textit{Eiffel Tower}; during the processing of social media conversations, a parallel analysis from wikis can help ML models understand that the event is closely associated with \textit{France}, \textit{Paris}, \textit{a structure}, \textit{a monument}, etc. 

\subsection{Scope and organization}

There are existing surveys related to social media analytics concerning its techniques, tools, platforms~\cite{batrinca2015social}, and applications---disasters~\cite{nazer2017intelligent,imran2015processing}, visual analytics~\cite{wu2016survey}, politics~\cite{stieglitz2013social}, health~\cite{abbasi2014social}, business~\cite{holsapple2014business}, false information and rumor detection~\cite{kumar2018false,zubiaga2018detection}. Based on our literature search, the previous surveys either focus on social media in general and/or are application/analysis-specific. To the best of our knowledge, this survey is the first that studies the use of AI for extracting situation awareness from microblogs without limitations on both the application and analysis. As major contributions of this survey, we:

\renewcommand\labelitemi{---}
\begin{itemize}
    \item discuss the essential aspects of microblog analytics and present a high-level methodological view of the literature (in Section~\ref{background}),
    \item survey AI research that extract situational information from microblog data (in Section~\ref{survey}),
    \item provide a commentative review of the literature and present challenges and research directions in the domain (in Section~\ref{discussion}).
\end{itemize}

\section{Background and Preliminary Concepts}
\label{background}

In this section, we discuss high-level aspects of microblog analytics that are common across the surveyed literature. We start with a review of differing tasks, associated algorithms/methods, some advanced ML methods, and later provide a broad description of commonly used microblog data and its uses, followed by basic and advanced methods for representation learning. We refrain from \emph{en masse} citations to the literature in this section; a detailed literature breakdown is given in Section~\ref{survey}.

\subsection{Fundamental Machine Learning}
\label{modelling-analyses}
Many of the applications/problems described in the surveyed literature were addressed using fundamental applications of machine learning and/or applied statistics.

\subsubsection{Supervised learning problems}
In this set of problems, the task is to learn a function that maps an input to output based on previously recorded examples~\cite{bishop2006pattern}; therefore requiring labelled data for training purposes. Considering a set of $N$ previously recorded examples, \begin{math}\{(x_{1},y_{1}),\dotsc, (x_{n},y_{n})\}\end{math}, such that $x_{i}$ represents the feature vector of the $i$-th example and $y_{i}$ represents the corresponding output class, a supervised learning problem aims at modelling a function $f:X\rightarrow Y$, where $X$ represents the input space and $Y$ represents the output space. Supervised learning problems are further divided into two types: \textit{Regression} (continuous output) and \textit{Classification} (discrete output or categories). The literature employs different variants of regression-based algorithms such as Linear/Polynomial Regression, Stepwise Regression, Poisson Regression \cite{wang2012automatic, bendler2014investigating, corea2016can, leitch2017twitter}, and classification-based algorithms such as Logistic Regression (LR) \cite{gerber2014predicting, chen2015crime, ashktorab2014tweedr}, Naive Bayes (NB) \cite{wang2012system, d2015real, gu2016twitter, sarker2016social, neppalli2017sentiment, ruz2020sentiment}, {$K$}-Nearest Neighbours (KNN) \cite{abbass2020framework}, Decision Trees (DT) \cite{d2015real, sarker2016social}, Random Forest (RF) \cite{alam2018twitter, ruz2020sentiment, tejaswin2015tweeting, mamidi2019identifying}, Support Vector Machine (SVM) \cite{d2015real, sarker2016social, neppalli2017sentiment, ruz2020sentiment}, and Artificial Neural Network (ANN) (refer to Section \ref{deep-trend} for recent trends in the use of neural models).

\subsubsection{Unsupervised learning problems}
This set of problems deal with the task of learning internal patterns from unlabelled data~\cite{bishop2006pattern}: \textit{clustering}---finding groups of similar examples; \textit{density estimation}---determining the distribution of data; \textit{visualization}---projecting high-dimensional data to a two or three-dimensional space; \textit{outlier detection}---detecting anomalies; \textit{topic modelling}---learning latent variables; and understanding opinions from texts using lexicon-based methods (some literature classify lexicon-based methods as semi-supervised as these methods maintain a predefined dictionary of words and phrases). Commonly used methods in unsupervised learning are $K$-Means \cite{liu2015social}, Hierarchical Clustering, Isolation Forest, Latent Dirichlet Allocation (LDA) \cite{gerber2014predicting, wang2012automatic, abd2020top}, Kernel Density Estimation (KDE) \cite{gerber2014predicting, chen2015crime}, Principal Component Analysis (PCA), and ANN. There is a third learning paradigm, an instance of \textit{weak supervision}~\cite{zhou2018brief}, known as \textit{Semi-supervised learning}, where a small set of labelled data is used alongside a large set of unlabelled data during training. One of its cases is \textit{Active learning} (discussed shortly in Section~\ref{active-learning}), which aims at training models during situations when labelled data may not be available promptly.

\subsubsection{Applied statistical problems}
Besides the learning problems discussed above, some surveyed literature uses computationally intensive applied statistical methods. Examples include classical time series forecasting methods \cite{achrekar2011predicting, gaurav2013leveraging} (e.g. auto-regressive (AR) models, moving average (MA) models, and their variants) for forecasting the future using historical data, Markov models \cite{barbera2015birds} for modelling systems with pseudo-random characteristics, resampling methods for estimating the precision of statistics, generalized additive models \cite{wang2012spatio, de2017dengue} for modelling interpretable predictor functions, the Granger causality test for causality analysis \cite{bollen2011twitter, zhang2012predicting} in multiple time series data.

\subsection{Advanced Machine Learning}
Of the surveyed literature, the more recent works tend to utilize an ensemble of techniques for focusing fundamentally on the ``learning'' perspective to deal with large-scale data, discussed below.

\subsubsection{Representation learning} Also known as \textit{feature learning}, representation learning~\cite{bengio2013representation} allows a system to automatically learn meaningful and useful representations of data. This set of techniques replaces traditional manual feature engineering, thus enabling systems to learn features by themselves and train for specific tasks. Examples include ANN, Sparse Coding, Independent Component Analysis, and Clustering methods. \textbf{\textit{Deep learning}}~\cite{lecun2015deep} is a representation learning method based on ANN, used for both supervised and unsupervised tasks, where each layer in the network learns to transform the input data into a more abstract and composite representation. Nowadays, deep learning architectures are the go-to methods for dealing with large-scale data, as these architectures often capture sophisticated non-linear relationships and outperform their traditional counterparts in numerous tasks further in. Deep learning architectures such as Multilayer Perceptron (MLP), Convolutional Neural Network (CNN), and Recurrent Neural Network (RNN) are supervised models. Architectures such as Self-Organizing Maps (SOMs), Boltzmann Machines, and AutoEncoders are unsupervised models.

\subsubsection{Active learning}
\label{active-learning}
In Active learning~\cite{settles2009active}, a case of semi-supervised learning, a learning algorithm (learner) interactively queries a human to label sampled examples. Since the learner itself determines the examples for labelling, the number of labelled data needed in active learning is often less than the number required in normal supervised learning. The learner uses probabilistic methods for uncertainty sampling of examples from the unlabelled dataset. The uncertainty sampling approach includes strategies such as least confidence, margin sampling, and entropy sampling. Other methods include Query-By-Committee, Expected Error Reduction and Weighted Methods.

\subsubsection{Few-shot learning} In Few-shot learning (FSL)~\cite{wang2020generalizing} the training set contains a limited number of labelled examples. FSL is mainly applied to supervised tasks such as image classification, text classification, and object recognition. There is another instance of FSL---\textit{Few-shot reinforcement learning}---that aims to find a policy given only a few state-action pairs. The literature classifies FSL tasks into three categories: \textit{Data-based} where training data is augmented by prior knowledge; \textit{Model-based} where the hypothesis space is limited by prior knowledge; and \textit{Algorithm-based} where search methods in the hypothesis space are altered by prior knowledge.

\subsubsection{Transfer learning} In Transfer learning (TL)~\cite{zhuang2020comprehensive} knowledge is transferred across related source domains. Data labelling in the case of microblogs can be time-consuming, expensive, or unrealistic during events such as disasters and riots. Semi-supervised learning does utilize a small chunk of labelled data alongside a large unlabelled set for improving learning accuracy. However, obtaining unlabelled data itself can be difficult in many cases. During such a scenario, TL has been reported to show promising results. In TL, a base network is trained for a base task using a dataset, and then the learned features are transferred to a different network for training on a target dataset concerning a target task~\cite{yosinski2014transferable}. TL works if the features are generic to both base and target tasks. Some examples of TL include the Inception model~\cite{szegedy2016rethinking} and ResNet model~\cite{he2016deep} for image data, and Word2vec~\cite{mikolov2013efficient} and GloVe~\cite{pennington2014glove} for text data.

\subsection{Microblog data}

\begin{figure}[t]
  \centering
  \includegraphics[width=\linewidth]{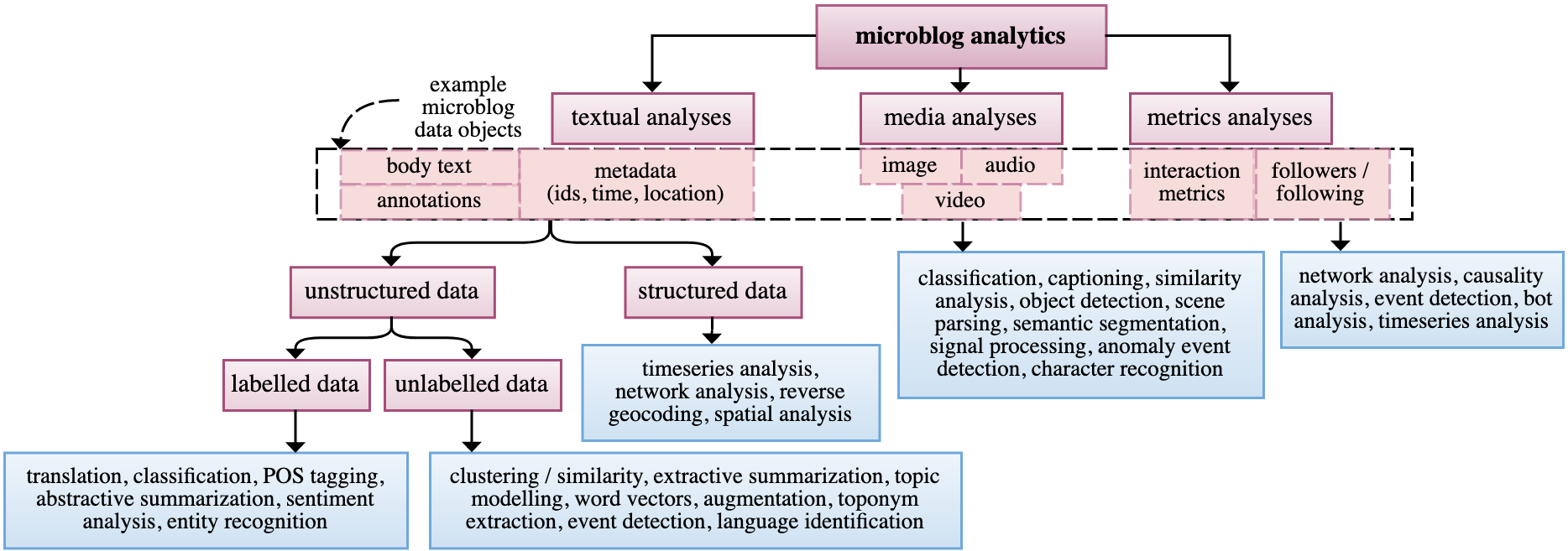}
  \caption{Taxonomy of \textit{microblog data objects}-specific tasks.}
  \label{taxonomy}
\end{figure}

The most common surveyed uses of microblog data are shown in Figure~\ref{taxonomy} as a taxonomy of \textit{microblog data objects}-specific tasks. From the literature, we have identified five main categories of microblog data: \textit{Data objects} include fundamental attributes such as text content, content creation time, content identifier, a resharing flag, etc.; \textit{User objects} include details that reference a user's profile, for example, user id, user biography, location, followers/following list, etc.; \textit{Geo objects}, available when content is geo-tagged, include attributes related to geographic information, such as precise geo-coordinates, bounding box coordinates, and place type; \textit{Entity objects} provide additional contextual information regarding the content, for example hashtags, URLs, user mentions, media, etc.; and \textit{Miscellaneous objects} contain diverse attributes, including interaction metrics, machine-identified language of the content and service specific data.

\subsection{Word Vectorization}
\label{words-to-vectors}
%Text data is typically pre-processed before its vector representation is computed. The text pre-processing task is use-case dependent and, in general, includes the following steps: (i) cleaning of symbols such as \texttt{``\#''} and \texttt{``@''}, URLs, extra spaces, indentation, line breaks, punctuation, and numbers, (ii) expansion of elements such as emojis to their text forms and abbreviations to their original forms, (iii) stemming/lemmatization for conversion of words to their root forms, (iv) lower casing, and (v) correction of incorrectly typed words. 

Word vector representation methods found in the literature are broadly either \emph{Frequency-based}, where the components of the vectors associate to the frequency attribute of words and word sequences, or \emph{Prediction-based}, where the vectors are trained weights of a neural network.

\subsubsection{Frequency-based}

\textbf{(a) Bag of words.}
In the Bag of words (BoW) model, each document (text object) is represented by a vector of dimension $|V|$ where $V$ is the vocabulary, and the $i$-th component of the vector represents the frequency of the corresponding $i$-th word in the document. \textbf{(b) Bag of $n$-grams.}
%The BoW model treats words as independent entities, thus failing to understand the context of words. The Bag of $n$-grams model exploits the intuition ``context originates from a sequence of words'' to represent a document as a bag of continuous word sequences. 
In this model, the vocabulary consists of $n$-grams, where $n$ represents the size of the continuous sequence. The $n$-gram of size $2$ is a \textit{bigram} and size $3$ is a \textit{trigram}. The BoW model is a special case where $n=1$, i.e. a \textit{unigram}. \textbf{(c) Term Frequency-Inverse Document Frequency (TF-IDF).} The TF-IDF model uses the ``importance of a word to a document'' factor for computing the vector representation of the document. The TF-IDF value, defined as the product of $TF(w,d) = frequency_{d}(w)$ and $IDF(w, D) = \ln \tfrac{|D|}{1+|{d\epsilon D:w\epsilon d}|}$, increases proportionally to the frequency of a word $w$ in a document $d$ and is penalized based on the number of documents in the corpus $D$ containing the word $w$. Frequency-based methods introduce high dimensionality and sparsity issues due to vocabulary size. Furthermore, they fail to capture the true semantics of words and do not deal with out-of-vocabulary (OOV) words.

\subsubsection{Prediction-based}
\textbf{(a) Static Embeddings.} In 2013, \citeauthor{mikolov2013efficient}~\cite{mikolov2013efficient} proposed \textbf{Word2vec}, a neural network-based text representation model that learns word vectors of a given dimension while maintaining semantic similarity between the words, thus solving the dimensionality, sparsity and context issues associated with frequency-based methods. %Word2vec uses two approaches for learning word vectors from large text corpus: the \textit{Continuous Bag-of-Words Model (CBOW)} for predicting the current word based on the context (surrounding words), and the \textit{Continuous Skip-gram Model} for predicting the context based on the current word.
\textbf{GloVe}~\cite{pennington2014glove} is a similar technique, but exploits both global statistics and local statistics of a text corpus, whereas Word2vec uses only the local statistics. Both Word2vec and GloVe consider words as the smallest units to train on. This results in an inability to deal with OOV words. In 2016, Facebook proposed \textbf{FastText}~\cite{bojanowski2017enriching} to address the OOV words issue by representing each word as a bag of character \textit{n}-grams. These models generate static embeddings (vectors) i.e. the same embedding is assigned to the same words used across different contexts.
%Therefore, contextualized word embeddings models were introduced for capturing word semantics in varying contexts.

\textbf{(b) Contextual (Dynamic) Embeddings.} Recent progress in Natural Language Processing (NLP) has led to the design of contextualized embedding models that address the issue of polysemous. Some widely used such models include seq2seq NMT (Neural machine translation) model-based \textit{Contextualized Word Vectors} (\textbf{CoVe}) \cite{mccann2017learned}; two-layer bidirectional LSTM-based \textit{Embeddings from Language Models} (\textbf{ELMo}) \cite{peters2018}; ASGD Weight-Dropped LSTM-based \textit{Universal Language Model Fine-tuning} (\textbf{ULMFiT}) \cite{howard2018universal}; and Transformer-based \cite{vaswani2017attention} \textit{Bidirectional Encoder Representations from Transformers} (\textbf{BERT})~\cite{devlin2018bert} and \textit{Generative Pre-trained Transformer-3} (\textbf{GPT-3})~\cite{radford2018improving}.

%Both models implement a shallow neural network, with non-linearity only in the output layer via the Softmax function. The size of the hidden layer is a hyper-parameter that defines the dimension of the word vectors to be learned. Once the training is completed, to obtain the word vectors (a.k.a. word embeddings), the weight matrix between the hidden layer and the output layer is considered in the case of the \textit{CBOW model}, and the weight matrix between the input layer and the hidden layer is considered in the case of the \textit{skip-gram model}.

%\begin{figure}[t]
%  \centering
%  \includegraphics[width=0.70\linewidth]{word2vec.png}
%  \caption{The two approaches used by the Word2vec model~\cite{mikolov2013efficient}. In the figure, $V$ is size of vocabulary; $C$ is the number of context words; $N$ is the word vector dimension; $W$ and $W'$ represent weight matrices.}
%  \label{word2vec}
%\end{figure}

\subsection{Image Representation}
\label{images-to-vectors}
In recent years, numerous methods have been proposed for learning image representations. Some methods to visual feature learning include learning through image colorization, image inpainting, image super-resolution, image generation with Generative Adversarial Networks (GANs) \cite{goodfellow2014generative} and game engines, context similarity and spatial relations between image patches, shuffling of image patches, image clustering, and TL \cite{jing2020self}. In 2020, \citeauthor{chen2020simple} \cite{chen2020simple} proposed \textit{SimCLR}, a simplified framework for contrastive learning of visual representations, that outperforms the existing state-of-the-art method \cite{henaff2020data} for self-supervised and semi-supervised learning, and performs on par with or better than the supervised baseline method \cite{kornblith2019better}.

%Images are also pre-processed depending upon their use-case before their learning their representations. Image data is generally: (i) resized to a fixed dimension, (ii) de-noised to remove ``outlier'' pixels for smoother color transition between edges, (iii) augmented by re-scaling and cropping, mirroring horizontally and vertically, and distorting pixel values associated with intensity and color, (iv) centered by subtracting the per-channel mean pixel, and (v) segmented (locate objects and boundaries) for more meaningful representation. An image is an array of pixel values with one or multiple color channels. A grey scale image has just one channel, while an RGB image has three channels: Red, Green, and Blue. A single-channel image can be represented by a two-dimensional array; however, an image with multiple channels requires an additional dimension, thus bringing \textit{tensors} into action. A tensor built upon the extracted pixel intensities of an image is then unrolled to form a vectorization of the image.

\subsection{Graph Representation}
\label{to-graphs}

Based on the encoder-decoder framework proposed by \citeauthor{hamilton2017representation} \cite{hamilton2017representation}, and depending on the type of information used in the encoder network, graphs representation learning methods are majorly classified into four classes: \textit{shallow embeddings}, \textit{graph auto-encoders}, \textit{graph-based regularization}, and \textit{graph neural networks} \cite{chami2020machine}. Shallow embeddings, auto-encoders, and graph neural networks are unsupervised embedding methods that map a graph onto a vector space with the objective of learning an embedding that preserves the graph structure. Graph-based regularization is a supervised graph embedding method; shallow embeddings and graph neural networks also have their use in supervised settings. The supervised methods, however, besides learning the representations, also predict node or graph labels that could be valuable particularly for downstream supervised tasks such as node or graph classification.

Mining of microblog data objects introduces various social networks of unprecedented scales: networks of friends, hashtags, keywords, followers, bots, geographic locations, etc. The literature represents a social network as a \textit{graph}---a nonlinear data structure consisting of entities and their relationships. Normally, a network is a graph $G(V, E)$, where $V$ denotes vertexes (entities) and $E$ denotes edges (relationships), and it is represented mathematically using matrices such as \textit{Incidence Matrix}, (Weighted) \textit{Adjacency Matrix}, \textit{Degree Matrix}, \textit{Laplacian Matrix} \cite{bollobas1998modern}. Social networks are a means to study the behavior of entities like people, organizations, and events; for example, a network analysis of users' pool is modeled by distinguishing the individuals as nodes and their relationships as edges. Graph mining has application to social network analysis across multiple tasks \cite{tang2010graph} including \textit{Centrality analysis}---identifying the most influential nodes in a network, \textit{Community detection}---identifying communities/groups through public discourse and/or interaction patterns, \textit{Information diffusion}---studying how information propagates/flows in a network, and \textit{Outlier detection}---identifying rare entities and/or relationships.

\subsection{Evaluation metrics}

There is no consistent use of evaluation metrics across the surveyed literature and this makes it hard to numerically compare different studies. R-squared ($R^{2}$), Mean Absolute Error (MAE), Root Mean Square Error (RMSE), Relative Absolute Error (RAE), and Root Relative Squared Error (RRSE) are some common evaluation metrics for regression problems. Consider a problem where $a_{1},\dotsc,a_{n}$ represents actual values and $p_{1},\dotsc,p_{n}$ represents predicted target values. Let $\epsilon=\sum(a_{i}-p_{i})^{2}$ and the notation $\overline{x}\equiv\tfrac{1}{n}\sum x_i$. Then the evaluation metrics are defined: $R^{2}$ is $1-\tfrac{\epsilon}{\sum (a_{i}-\overline{p})^{2} }$; \textit{MAE} is $\tfrac{1}{n}\sum|p_{i}-a_{i}|$; \textit{RMSE} is $\sqrt{\tfrac{\epsilon}{n}}$; \textit{RAE} is $\tfrac{\sum|p_{i}-a_{i}|}{\sum|a_{i}-\overline{a}|}$; and \textit{RRSE} is $\sqrt{\tfrac{\epsilon}{\sum(a_{i}-\overline{a})^{2}}}$. Some studies use Pearson's Correlation Coefficient (PCC), $r=\sfrac{\sum (x_{i}-\overline{x})(y_{i}-\overline{y})}{\sqrt{\sum (x_{i}-\overline{x})^{2} \sum (y_{i}-\overline{y})^{2}}}$ where $n$ is sample size, and $x_{i}$, $y_{i}$ are data points, to find a measure of fit for linear regression, and time series problems.

\textit{Accuracy}, \textit{Precision}, \textit{Recall} (a.k.a. \textit{sensitivity}), \textit{Specificity}, \textit{F-measure}, and \textit{Area under the ROC Curve} (AUC) are the commonly used evaluation metrics for classification problems. True Positive (TP), True Negative (TN), False Positive (FP) and False Negative (FN) are measures used for computing those metrics. TP represents the number of samples predicted positive and that are actually positive. TN represents the number of samples predicted negative and that are actually negative. FP represents the number of samples predicted positive but that are actually negative. FN represents the number of samples predicted negative but that are actually positive. Accuracy is the number of correct predictions to the total number of samples, $\tfrac{TP+TN}{TP+TN+FP+FN}$. Accuracy is not applicable if the dataset is imbalanced (unequal distribution of samples between classes). Precision gives the percentage of positive samples out of the total predicted positive samples and is computed as $\tfrac{TP}{TP+FP}$. Recall gives the percentage of positive samples out of the total actual positive samples and is computed as $\tfrac{TP}{TP+FN}$. Specificity gives the percentage of negative samples out of the total actual negative samples and is computed as $\tfrac{TN}{TN+FP}$. F-measure is the harmonic mean of precision and recall and is computed as $F_{\beta}=\frac{(1+\beta^{2})\times precision \times recall}{(\beta^{2} \times precision) + recall}$. The Area under the ROC Curve (AUC) is a plot of \textit{Sensitivity} versus \textit{1-Specificity} across varying threshold values. The Receiver Operating Characteristics (ROC) curve is a probability curve. A model with a good measure of separability has an AUC score near $1$.

Common evaluation metrics for clustering are \textit{Silhouette Coefficient} (SC) and \textit{Dunn’s Index} (DI). \textit{SC} is computed as $s=\frac{b-a}{max(a,b)}$, where $a$ is the average distance between a sample and all other datapoints in the same cluster, and $b$ is the distance between a sample and all other datapoints in the next nearest cluster. The distance metric can be any distance function, such as Euclidean or Manhattan. DI is the ratio of the minimum inter-cluster distance to the maximum cluster size.

%Text summarization models are typically evaluated using the \textit{Recall-Oriented Understudy for Gisting Evaluation} (ROUGE) metric. ROUGE-$N$ refers to the overlap of $n$-grams between generated and reference summaries.

\subsection{A high-level methodological view of the literature}
\label{pipeline}

\begin{figure}
  \centering
  \includegraphics[width=0.75\linewidth, height=10cm]{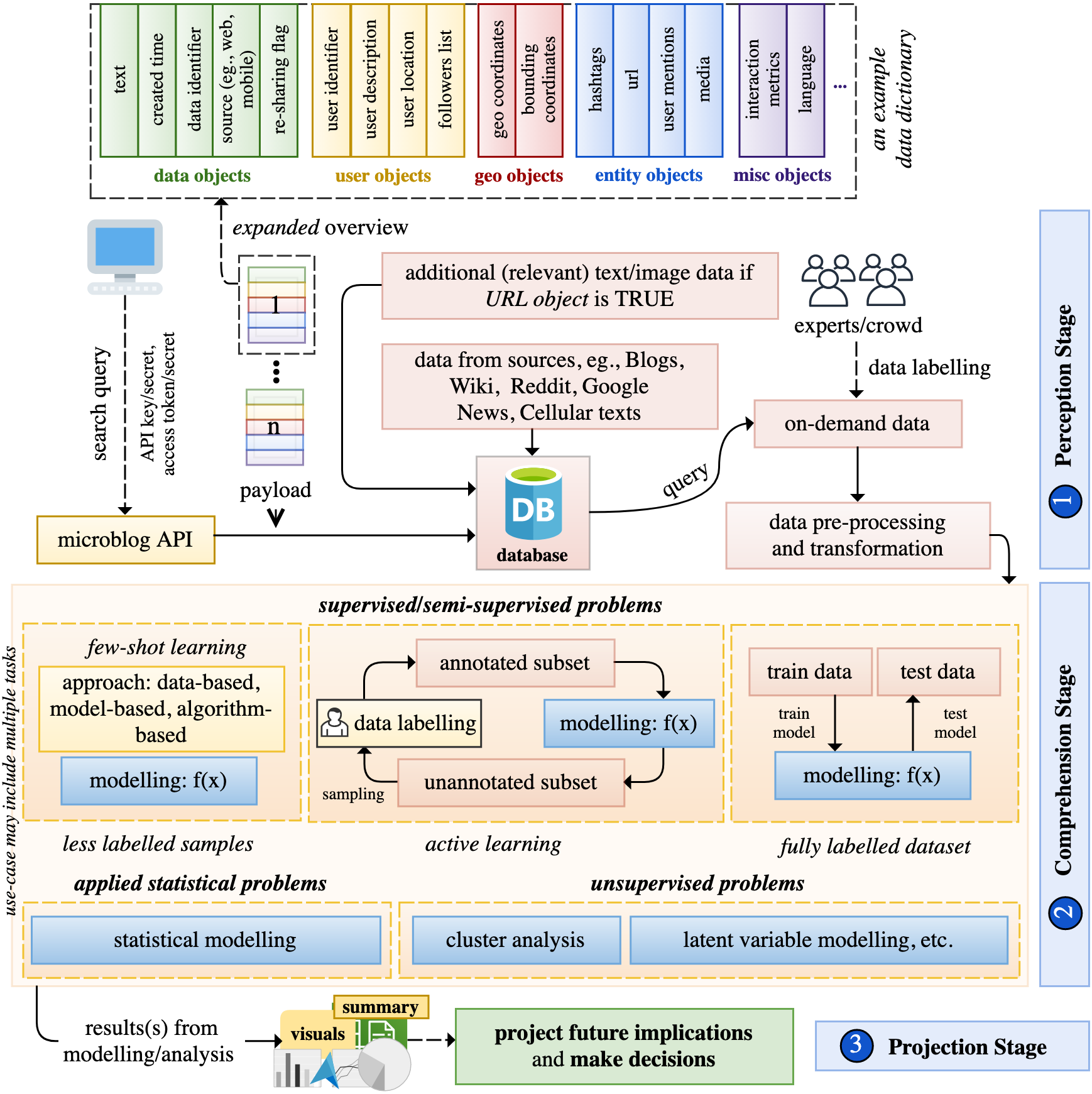}
  \caption{A high-level methodological view of the literature}
  \label{data-pipeline}
\end{figure}

To better understand how the surveyed literature relates to the common problem of enhancing situation awareness from microblog data, we present a high-level methodological view of the literature, shown in Figure~\ref{data-pipeline}, that embodies the microblog data objects and range of techniques discussed above, and we provide theme-specific methodological examples in Section~\ref{survey}.

\subsubsection{The perception stage}
This stage involves the collection of relevant data. The first step to curating a large-scale microblog dataset starts with pre-determining a set of ``seed'' \textit{keywords} (words and phrases) and \textit{hashtags} (keywords preceded by a hash sign). Next, a \textit{search query} is created. A search query contains keywords, hashtags, and conditions for requesting relevant data from API endpoints. For example, a search query using \textit{twarc}\footnote{https://github.com/DocNow/twarc} on Twitter's \textit{full-archive endpoint} can be as simple as: \textit{"(covid OR quarantine OR lockdown OR pandemic OR \#covid-19 OR \#covid OR mask OR ppe OR flu) -filter:nativeretweets lang:en" -{}-archive -{}-since-id 1372953264359145481 -{}-until-id 1407693866816393223}. This query searches for original (not re-shared) English language Twitter contents that include the queried keywords and hashtags within their text body and were created between the two time-based ``Snowflake'' identifiers. When a connection is established, the microblog API endpoint returns a continuous data stream---\textit{payload}---based on the search query. The payload is stored in a central database. Additional data such as discussions, headlines, and reports from other sources can also be considered for enriching the existing corpora.

High volume endpoints such as Twitter's \textit{COVID-19 stream endpoint} utilizes 4 partitions to split the overall volume of payload over multiple connections to consume the entire stream. File-based databases such as \textit{SQLite} are equally effective for handling large-scale data, but they come with demerits of their own compared to their distributed peers. Therefore, the selection of the database is up to the user. Next an on-demand dataset is created and labelled (if the problem is supervised) either by experts or through crowdsourcing. The resulting data is finally pre-processed as per requirements.

\subsubsection{The comprehension stage}
This stage deals with data modelling and analyses and  is problem-specific (refer to Section \ref{modelling-analyses}) and relates to tasks such as Text Classification, Text Clustering, Image Classification, Sequence Labelling, etc. Refer to Figure \ref{taxonomy} for a list of microblog objects and their corresponding tasks.

\subsubsection{The projection stage}
This stage involves observing visualizations and summaries generated during the comprehension stage for projecting current scenarios to their future implications and making decisions. Activities in this stage include: (i) observing abstractive summarization of the data stream and quantitative assessments on maps/charts, and (ii) analyzing real-time convex closures (generated using geo-tagged content) of a situation.

\section{Survey of AI Approaches for Situation Awareness}
\label{survey}
\begin{figure}[t]
  \centering
  \includegraphics[width=0.7\linewidth, height=8.5cm]{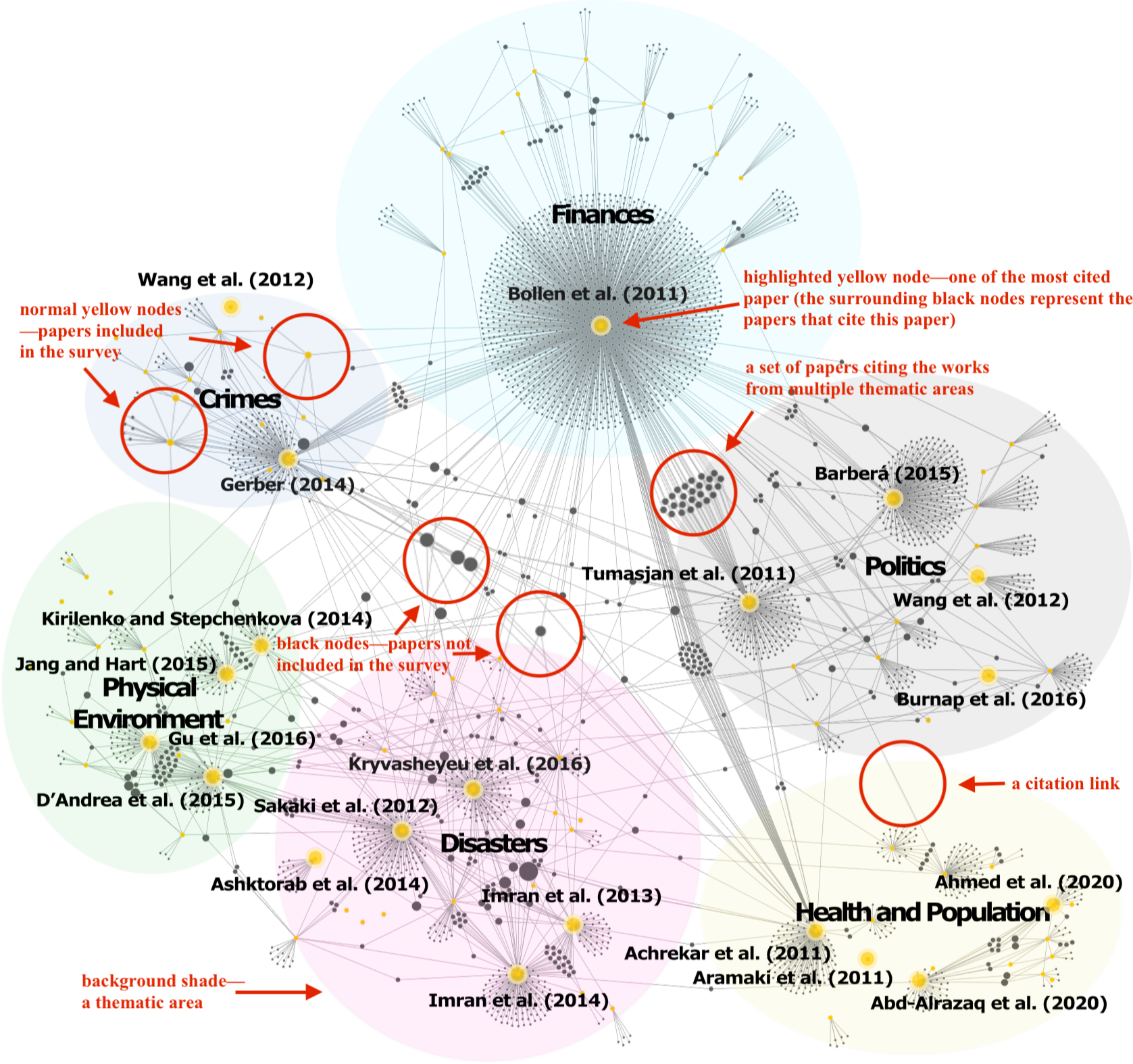}
  \caption{The citation network. Yellow nodes represent papers in the survey, manually clustered into thematic area, and black nodes represent papers that cite the surveyed papers. The 20 most cited papers are labelled. \textit{Some nodes have irregularities due to limitations at the API end.}}
    \label{citation-network}
\end{figure}

We undertook an extensive search in the digital libraries of \textit{ACM}, \textit{Elsevier}, \textit{IEEE}, \textit{Springer}, and other major publishers while also utilizing the \textit{Web of Science} and \textit{Scopus} databases to identify relevant papers published between 2010--21 (the early 2010s mark the emergence of microblog APIs\footnote{https://blog.twitter.com/developer/en\_us/a/2011/streaming-api-turning-ssl-only-september-29th}). We discarded papers from the initial set if: (i) primary data source was not microblog, (ii) extracting ``situational information'' was not the primary objective, (iii) AI was not employed significantly, and (iv) published before 2020 and had <5 citations. \textit{Crossref}'s public API\footnote{https://www.crossref.org/documentation/retrieve-metadata/rest-api/} was used for querying the list of papers that cited a set of 78 papers initially collected during the filtering stage of the paper selection process. This data was used to create a citation network, shown in Figure~\ref{citation-network}, where nodes represent papers and are weighted based on the number of citations they receive from within the network. Under the hypothesis that the most influential papers are likely to have many citations we manually identified 19 additional papers that were either foundational papers or significantly related to the scope of this survey. Out of the surveyed papers, the nodes corresponding to the most prominent papers in terms of citations, are highlighted and labelled in the citation network. 

\textbf{Literature breakdown.} We reviewed 97 papers out of which 28 were from \textit{Elsevier}, 15 from \textit{IEEE}, 13 from \textit{ACM}, 9 from \textit{JMIR}, 6 from \textit{Springer}, and 26 (combined) from journals such as \textit{Taylor \& Francis}, \textit{Wiley}, \textit{SAGE}, \textit{Emerald} and \textit{BMC}. Out of 97 papers, this survey involved 63 journal papers, 29 conference and symposium papers, 2 workshop papers, 2 book chapters and 1 thesis article. Among the 20 most cited papers, 15 of them were published before 2016. Out of those, 7 are journal papers, 6 are conference papers, 1 is a book chapter, and 1 is a workshop paper. The remaining 5 out of the 20 most cited papers published after 2015 are all journal articles.

%Figure \ref{papers-dist-type} shows the distribution of papers based on venue type, and Figure \ref{papers-dist-year-type} shows a more granular distribution concerning year and paper type.

%\begin{figure}[t]
%\begin{subfigure}{.45\textwidth}
%  \centering
%  \includegraphics[width=1\linewidth]{papers-dist-type.png}
%  \caption{Based on venue type}
%  \label{papers-dist-type}
%\end{subfigure}%
%\begin{subfigure}{.55\textwidth}
%  \centering
%  \includegraphics[width=1\linewidth]{papers-dist-year-venue.png}
%  \caption{Based on year and venue type (2021 is an ongoing year)}
%  \label{papers-dist-year-type}
%\end{subfigure}
%\caption{Distribution of papers considered in the survey}
%\label{papers-dist}
%\end{figure}

\textbf{Thematic areas.} The papers identified for the survey are classified into six thematic areas: Crime, Disasters, Finance, Physical Environment, Politics, and Health and Population. Each thematic area is further divided into sub-thematic areas based on research directions. Tables~\ref{lit-crimes}--\ref{lit-health} provide an overview of the essential elements of the literature in each thematic area, and we use abbreviations throughout as shown in Table~\ref{abbreviation}.
\begin{scriptsize}
\begin{table}[t]
  \caption{Abbreviations used in Tables~\ref{lit-crimes}--\ref{lit-health}}
  \label{abbreviation}
  \begin{tabular}{p{0.9cm}p{4cm} p{0.9cm}p{2cm} p{0.9cm}p{2.9cm}}
    \toprule
     
\textit{SmPar} & Semantic Parsing & \textit{TpMd} & Topic Modelling & \textit{SmEx} & Semantic Extraction\\

\textit{SpTmMd} & Spatio-temporal Modelling & \textit{Cf} & Classification & \textit{SeqLbl} & Sequence Labelling\\

\textit{Reg} & Regression & \textit{SnAn} & Sentiment Analysis & \textit{WoVec} & Word Vectors\\

\textit{DnEs} & Density Estimation & \textit{AdSt} & Advanced Statistics & \textit{CoFqAn} & Content Frequency Analysis \\

\textit{LLVFE} & Low Level Visual Features Extraction & \textit{NetAn} & Network Analysis & \textit{NER} & Name Entity Recognition\\ 

  \bottomrule
\end{tabular}
\end{table}
\end{scriptsize}

\subsection{Crime}
This thematic area surveys microblog data as a ``social crime sensor'' that helps detect geographical regions that are more likely to show criminal conduct. Such analysis can assist agencies and governments in producing persistent solutions to counter crimes such as theft, burglary, robbery, cyberbullying, and harassment. Table~\ref{lit-crimes} provides an overview of the literature in this theme.

\subsubsection{Predicting future crime}
A Generalized Linear Regression Model (GLM) was designed by \citeauthor{wang2012automatic}~\cite{wang2012automatic} for predicting future hit-and-run incidents. The authors used LDA for topic modelling and Semantic Role Labeling (SRL) for extracting events, entities and their relationships from tweets. Their results showed that training an LDA/GLM model on event-specific words has higher predictive capability than training the same model on a complete vocabulary. \citeauthor{wang2012spatio}~\cite{wang2012spatio} extended this work while incorporating spatio-temporal, geographic, and demographic data. The extended work proposed a new modelling approach, the Spatio-Temporal Generalized Additive Model (STGAM), that used a feature-based approach to predict the probability of criminal activities alongside their space and time attributes.

\citeauthor{gerber2014predicting}~\cite{gerber2014predicting} collected geo-tagged Twitter data, used LDA for topic modelling, and designed a binary LR model using KDE + the features derived from the Twitter topics. In 19 of the 25 crime types, the Twitter data + KDE features improved the model's performance compared to using just the KDE features. Similarly, \citeauthor{chen2015crime}~\cite{chen2015crime} considered crime density, the sentiment of tweets, the last three days' sentiment trend, and weather factors such as temperature, dew, precipitation, etc., as explanatory variables for designing a model to predict future crimes. The authors used KDE to compute crime density and lexicon-based methods for sentiment analysis and reported the KDE + Twitter + weather data model surpassing the base model that used only KDE features.

\citeauthor{bendler2014investigating}~\cite{bendler2014investigating} used predictive analytics to show that Twitter data can assist in predicting future crimes. The authors selected a small region within San Francisco with higher Twitter activity, divided the area chosen into a $10\times 10$ grid, and performed regression analysis to provide evidence for the relationship between Twitter data and criminal activities. They trained SVM models to predict burglary and robbery crimes while including and excluding tweets' volume feature. The inclusion of volume feature for training showed improvement in prediction performance.

Foursquare data has also been used in combination with Twitter data to study the correlation between the predicted masses of people at different venues and the occurrence of real crimes at those venues. \citeauthor{wang2015using}~\cite{wang2015using} designed two text-enriched models, one for predicting the type of venue (e.g., restaurant or transportation hub) the user is likely to visit next and another for predicting the spatial information. Their models outperformed all baseline models---three Markov models and an SVM model---trained on historical visiting details. They reported that geo-tagged tweets do correlate with users' next venue visits.

\begin{scriptsize}
\begin{table}[t]
  \caption{Overview of the literature in ``Crime'' thematic area}
  \label{lit-crimes}
  \begin{tabular}{p{1.2cm} l l p{2.5cm} p{1.35cm} p{2.3cm} p{3cm}}
    \toprule
    \textbf{Direction} & \textbf{Year} & \textbf{Study} & \textbf{Primary Dataset} & \textbf{Geo Scope} & \textbf{Tasks}  & \textbf{Best outcome}\\
    \midrule
\multirow{10}{1cm}{Predicting crimes} & 2012 & \cite{wang2012automatic} & 3.65k tweets &  United States & \textit{SmPar}, \textit{TpMd} & Wide confidence intervals\\

 & 2012  & \cite{wang2012spatio} & same as \cite{wang2012automatic} & United States & \textit{SmPar}, \textit{SpTmMd}, \textit{TpMd}  & AUC: 0.7616\\

 & 2014 & \cite{bendler2014investigating} & 60k geo-tagged tweets & United States & \textit{Cf}, \textit{Reg}  & Accuracy: 0.66\\

 & 2014 & \cite{gerber2014predicting} & 1.52k geo-tagged tweets & United States & \textit{DnEs}, \textit{TpMd}  & AUC: 0.71\\

 & 2015 & \cite{chen2015crime} & 1.06M tweets & United States & \textit{Cf}, \textit{DnEs}, \textit{SnAn}  & AUC: 0.67\\

 & 2015 & \cite{wang2015using} & 1.23M tweets, 224k POI from Foursquare & United States & \textit{Cf}, \textit{Reg} & Accuracy: 71\%\\

 & 2018 & \cite{vomfell2018improving} & 6 month period tweets, 47k POI from Foursquare  & United States & \textit{Reg} & 19\% improvement over baseline model\\

 & 2020 & \cite{abbass2020framework} & 150k tweets & Global & \textit{Cf}  & Accuracy: 92.0\%\\
\midrule

\multirow{3}{1cm}{Analyzing crime rates} & 2016 & \cite{aghababaei2016mining} & 101M tweets & United States & \textit{Cf} &	F-measure: 0.83\\

 & 2018 & \cite{aghababaei2018mining} & 1.13M tweets &  United States & \textit{Cf}, \textit{SnAn}, \textit{TpMd} & F-measure: 0.94\\

& 2020 & \cite{vo2020crime} & 3.80k tweets & India & \textit{Clu}  & Accuracy: 70\%\\
\midrule

\multirow{5}{1cm}{Space-time Analysis} & 2019 & \cite{pina2019exploring} & 26k tweets, 2.57k tweets after filtering & Mexico & \textit{AdSt}, \textit{Reg} &	Absolute errors for 80\% \& 95\% coverage: 3.796 \& 2.933\\

 & 2019 & \cite{siriaraya2019witnessing} & 979k geo-tagged tweets	& Mexico City & \textit{SnAn}, \textit{WoVec}  & F-measure: 0.80\\

 & 2020 & \cite{ristea2020spatial} & 9.43k geo-tagged tweets	& Mexico City & \textit{DeEs}, \textit{SnAn} & AUC upper limit: 0.72--0.77\\

 & 2021 & \cite{park2021happy} & 123k tweets & Mexico City & \textit{Clu}, \textit{SnAn}, \textit{TpMd} & Statistical analysis presented\\
  \bottomrule
\end{tabular}
\end{table}
\end{scriptsize}

Similarly, data from Foursquare and Twitter has been combined with taxi trip data to analyze human activity patterns. \citeauthor{vomfell2018improving}~\cite{vomfell2018improving} presented a multi-model solution using a Simultaneous Autoregressive Model, a Conditional Autoregressive Model, and a Generalized Linear Mixed Model as some of the predictors for their spatial linear regression models. The authors trained RF, Ensemble-based, and ANN models and reported that the heterogeneous data sources they considered contribute to a better prediction of property crimes compared to using just demographic data.

\citeauthor{abbass2020framework}~\cite{abbass2020framework} developed an $n$-gram language model to predict cybercrimes using Twitter. The authors collected tweets containing hashtags such as \texttt{\#harassment}, \texttt{\#metoo}, \texttt{\#sexualassualt}, \texttt{\#cyberbullying}, \texttt{\#victim} for training Multinomial NB, KNN and SVM models to find the best value of $n$ for the $n$-gram model. Their results showed that the bigram language model performed better than the network-based feature selection approach, and SVM outperformed other models.

\subsubsection{Analyzing crime rates}
Relationships between Twitter data and crime rates have been examined in multiple studies. \citeauthor{aghababaei2016mining}~\cite{aghababaei2016mining} trained a linear SVM model using features extracted from historical Twitter data. Their model achieved an $F$-measure of 0.83 for crimes such as theft, burglary, and sex offences; however, the results for crimes such as murder and vandalism correlated poorly. The same work was later extended with a temporal topic model~\cite{aghababaei2018mining} which outperformed the batch model in 17 out of 22 crime types. \citeauthor{vo2020crime}~\cite{vo2020crime} analyzed Twitter data from seven major cities of India to confirm that tweets contribute to a better understanding of crime rates. The authors used the Twitter part-of-speech tagger\footnote{http://www.cs.cmu.edu/~ark/TweetNLP/}, and a class-based $n$-gram clustering~\cite{brown1992class} to build a crime rate detection model that predicted nearly 70\% of the crime rates.

\subsubsection{Space-time analysis}
An exploratory study was done by \citeauthor{pina2019exploring}~\cite{pina2019exploring} using Twitter data alongside Google Trends on 13 different crimes, including theft, robbery, rape and homicide. Tweets were filtered at the API level using Spanish keywords such as \texttt{``inseguridad''}, \texttt{``violencia''}, \texttt{``robo''}, \texttt{``crimen''}, and \texttt{``v\'ictima''}. Their results showed the pairwise correlation of the official crime data and the tweets as almost negligible. The authors asserted that Twitter data should be considered merely observative and not representative. However, they acknowledged the effectiveness of Twitter data in understanding the spatio-temporal patterns of crime data.

A crime analysis tool was developed by \citeauthor{siriaraya2019witnessing}~\cite{siriaraya2019witnessing} to provide contextual information regarding crimes through visualization. The tool uses a binary linear SVM trained on GloVe for identifying negative sentiment tweets to explore the negative characteristics of crime associated areas. The tool visualizes situational information based on crime type and period. Tweets specific to a crime or a region are shown on an interactive map, and a word cloud is generated to describe the spatial and temporal aspects of the crime area.

Concerning sporting events, \citeauthor{ristea2020spatial}~\cite{ristea2020spatial} studied the spatial relationship between geo-tagged tweets and crime occurrences alongside demographic and environmental aspects. The authors used crowd-sourced databases related to ``hate words'' and ``swearing'' to extract tweets containing at least one word from those databases. Feature selection was done using RF, and sentiment analysis was performed based on lexicon methods. They computed the density estimate based on the centre points of each grid and then trained an LR model. Their result showed improved AUC in all crime types during event days and non-event days with the addition of Twitter data to the base model. 

Similarly, \citeauthor{park2021happy}~\cite{park2021happy} examined the spatial distribution of sentiments and the level of happiness using geo-tagged Twitter data, census data and geospatial data on one of the shrinking cities in the United States---Detroit. The authors collected username, text, geo-coordinates and used the Complete Automation Probability (CAP) technique to filter out bot-involved tweets. Cluster and hotspot analyses were performed to study the spatial distribution of sentiments, LDA was implemented for topic modelling, and the AFINN lexicon dictionary\footnote{https://pypi.org/project/afinn/} was used for computing sentiments of tweets. Their results reported that areas with less crime tend to be happier and suggested that areas with negative sentiments should be prioritized for regeneration efforts.

\begin{figure}[t]
  \centering
  \includegraphics[width=0.75\linewidth]{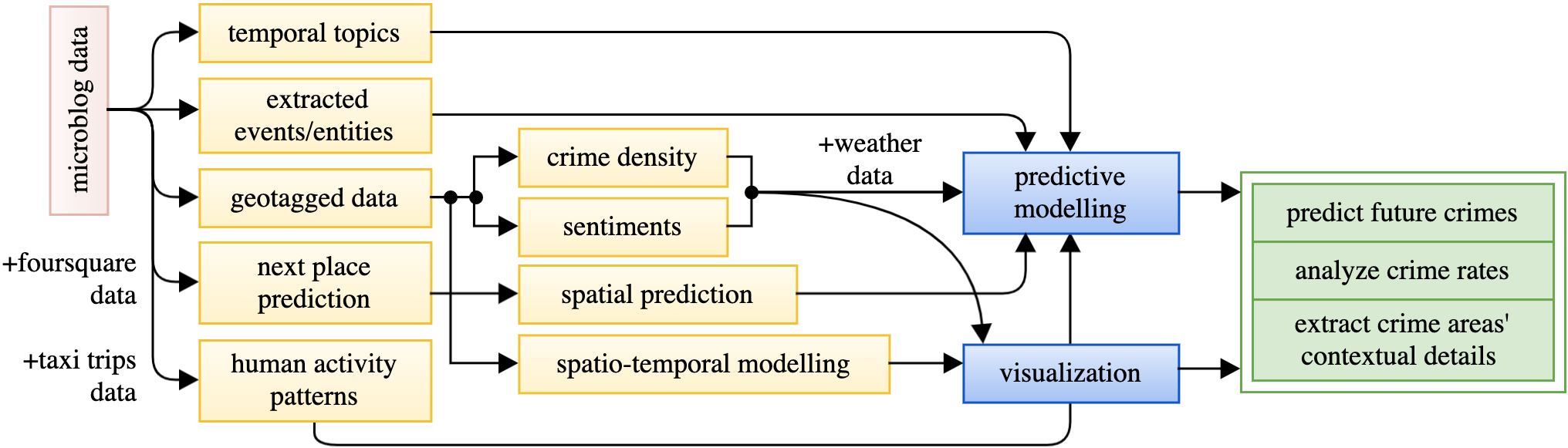}
  \caption{A methodological view of the literature in ``Crime'' thematic area}
  \label{crisis-mindmap}
\end{figure}

An illustrative summary of methodologies discussed in this thematic area is given in Figure~\ref{crisis-mindmap}.

\subsection{Disaster}
People disseminate situational information before, during, and after natural/human-made disasters and hazards such as earthquakes, floods, cyclones, fires, terrorist attacks, and nuclear explosions. Researchers have used such socially generated data for enhancement of situation awareness and prediction and rapid assessment of disasters with a common purpose to facilitate law enforcement agencies and humanitarian bodies to get firsthand knowledge about an ongoing situation. Table~\ref{lit-disasters} provides an overview of the literature in this theme.

\subsubsection{Identifying relevant data}
\citeauthor{imran2013practical}~\cite{imran2013practical} applied a classification-extraction approach, also presented in their previous work~\cite{imran2013extracting}, for extracting informative tweets. The authors implemented a sequence labelling algorithm to tokenize tweets to form a sequence of word tokens and assign labels based on classification requirements. The study intended to detect informative tweets and distinguish those tweets as reported by direct eyewitnesses or simply the repetitive blocks of earlier reported situations. The classification task was validated against tweets labelled by volunteers to show that their approach could extract 40-80\% of informative tweets.

A keyword relevance scoring approach was developed by \citeauthor{joseph2014approach}~\cite{joseph2014approach} for identifying and ranking keywords that might assist in finding tweets that describe a situation. The authors used reports generated on Ushahidi, a crowd-sourced platform, to identify keywords associated with actions, entities and location aspects of a disaster. The relevance of each keyword was computed based on its frequency in Ushahidi reports and its use in other contexts besides disaster.

A real-time multi-level damage assessment model was proposed by \citeauthor{shan2019disaster}~\cite{shan2019disaster}. In the first level, dictionaries specific to physical damage and sentiments were constructed. LDA was used on infrastructure, industry, people, etc., related documents retrieved from the Baidu search engine. In the second level, the situation (what), time (when) and entity (who) information gathered from Weibo posts were summarized and the physical dictionary was used for assessing the level of physical damages, and the sentiment dictionary for assessing emotional damage. In the third level, a semantic web was created to extract semantic patterns by analyzing prominent ``noun'' based nodes such as building, factory, etc. Similarly, \citeauthor{karami2020twitter}~\cite{karami2020twitter} proposed a Twitter data analysis framework, TwiSA, that implements lexicon-based sentiment analysis and LDA-based temporal topic discovery. The framework concentrates on text analysis for extracting situational information to support the preparedness, response and recovery phases of the disaster management cycle. 

A two-stage classification model was proposed by \citeauthor{rizk2019computationally}~\cite{rizk2019computationally} for identifying tweets that report damages to either human-made structures or the natural environment. First, visual features and semantic features were extracted independently from tweets. The extracted features were then passed onto the next stage of classifiers for the final classification. SVM, Ensemble-based and ANN models were trained on a disaster tweets corpus and a scene dataset. A domain-specific dense BoW model was generated for building semantic descriptors. The authors reported that the addition of semantic features alongside low-level visual features improved classification performance.

\citeauthor{zahra2020automatic}~\cite{zahra2020automatic} trained multiple classifiers for four different disasters---floods, hurricanes, earthquakes, and wildfires---using crowd-sourced labelled data to classify disaster specific tweets into three categories: eyewitness, non-eyewitness, and don't know. The authors used different combinations of textual features and domain-expert selected features. The study reported that perceptual senses are usually found in tweets created by direct eyewitnesses, whereas emotions and prayer specific terms are found in tweets from indirect eyewitnesses.

Similarly, \citeauthor{alam2018twitter}~\cite{alam2018twitter} performed a textual content analysis on Twitter data collected during three hurricanes---Harvey, Irma, and Maria. The study involved understanding humanitarian needs by using RF classifiers, learning the concerns of affected population through sentiment using Stanford's sentiment analysis classifier\footnote{https://nlp.stanford.edu/sentiment/}, tracking incidents by implementing LDA for topic-modelling, and identifying notable entities through name entity recognition. The authors also performed a multimedia analysis using a TL-based deep image recognition model to identify the relevance of images to disastrous events.

\begin{scriptsize}
\begin{table}
  \caption{Overview of the literature in ``Disaster'' thematic area}
  \label{lit-disasters}
  \begin{tabular}{p{1.3cm} p{0.4cm} p{0.4cm} p{2.35cm} p{2.65cm} p{1.9cm} p{2.4cm}}
    \toprule
    \textbf{Direction} & \textbf{Year} & \textbf{Study} & \textbf{Primary Dataset} & \textbf{Geo Scope} & \textbf{Tasks} & \textbf{Best outcome}\\
    \midrule
\multirow{8}{1cm}{Identifying relevant data} & 2013 & \cite{imran2013practical} & 346k tweets & Joplin$^{T}$, Sandy$^{H}$ &  \textit{Cf}, \textit{SeqLbl}  & Accuracy: 80--90\%\\

 & 2014 & \cite{joseph2014approach} & 1k Ushahidi reports, 90M tweets & Haiti &  \textit{AdSt} &	Qual. disc. presented\\
 
 & 2018 & \cite{alam2018twitter} & 9M tweets, 191k images & Harvey$^{H}$, Irma$^{H}$, Maria$^{H}$ & \textit{Cf}, \textit{NER}, \textit{TpMd} &	Accuracy: 76--90\%\\

 & 2019 & \cite{rizk2019computationally} & 1.34k tweets & Nepal, Chile, Japan, Kenya & \textit{Cf}, \textit{LLVFE}, \textit{SemExt} & Accuracy: 92.43\%\\
 
 & 2019 & \cite{shan2019disaster} & 67k Weibo posts & China & \textit{SmEx}, \textit{TpMd} & 5/8 close predictions\\

 & 2020 & \cite{zahra2020automatic} & 20M tweets & Global &	\textit{Cf} & F-measure: 0.69--0.95\\

 & 2020 &  \cite{karami2020twitter} & 1M tweets & South Carolina & \textit{SnAn}, \textit{TpMd} & Qual. disc. presented\\

\midrule
\multirow{4}{1cm}{Mining systems} & 2012 & \cite{sakaki2012tweet} & 597 tweets & Japan & \textit{Cf}, \textit{SemExt}, \textit{SnAn} &	F-measure: 0.73\\

& 2014 & \cite{imran2014aidr} & 200 tweets   & Global &	\textit{Cf} &	AUC: 80\%\\

 & 2014 & \cite{ilyas2014microfilters} & 700--850 images & Global &	\textit{Cf} & AUC: 78\%\\

& 2014 & \cite{ashktorab2014tweedr} & 17M tweets & North America & \textit{Cf}, \textit{Clu} &	AUC: 88\%\\
\midrule

\multirow{6}{1cm}{Assessment of disasters} & 2016 & \cite{kryvasheyeu2016rapid} & 52.5M & Sandy$^{H}$ & \textit{AdSt}, \textit{SnAn} &	Stat. analysis presented\\

& 2017 & \cite{neppalli2017sentiment} & 12.9M tweets &  Sandy$^{H}$ &	\textit{Cf}, \textit{SnAn} &	Accuracy: 75.91\%\\

& 2018  & \cite{resch2018combining} & 1.5k tweets (subset) & California &	\textit{TpMd} & Fleiss kappa score: 0.886\\

 & 2018 & \cite{wu2018disaster} & 970k tweets & Sandy$^{H}$ & \textit{AdSt}, \textit{SnAn}, \textit{TpMd} &	Corr. coef.: 0.989\\
 
 &  2019 & \cite{hernandez2019using} & 7.76k unique tweets & Mexico City & \textit{Cf}, \textit{DnEs} & Avg. F-measure: 0.84 \\

 &  2020 & \cite{zhai2020examine} & 39k geo tweets &  Florence$^{H}$ &	\textit{Cf}, \textit{TpMd} &	Avg. Accuracy: 74.23\\

\midrule

\multirow{4}{1cm}{Benchmarking}  & 2016 & \cite{caragea2016identifying} & 2.82k tweets & Global &	\textit{Cf} &	Accuracy: 75.90--82.52\%\\

& 2017  &\cite{nguyen2017robust} & 21k tweets &	Global & \textit{Cf} & AUC: 81.21--94.17\%\\

& 2018 & \cite{ragini2018big} & 70k tweets & India, Pakistan &	\textit{Cf} & F-measure: 0.72--0.94\\

 & 2020 &  \cite{ruz2020sentiment} & 2.18k/60k tweets & Chile, Catalonia &	\textit{Cf} &	Accuracy: 81.2--85.8\%\\

  \bottomrule
\end{tabular}\\
$^{T}$represents tornado.$^{H}$represents hurricane.
\end{table}
\end{scriptsize}

\subsubsection{Mining Systems}
\citeauthor{imran2014aidr}~\cite{imran2014aidr} presented a data mining platform, Artificial Intelligence for Disaster Response (AIDR), for the automatic classification of crisis-specific Twitter data into categories related to community needs, loss of lives and damages. The platform utilizes three core components: a collector for retrieving tweets, a tagger for classification of tweets, and a trainer for training the classifier present in the tagger component. The platform automatically collects tweets created during humanitarian crises and trains on a sample of labelled data. A similar classification platform with an SVM classifier, MicroFilters, was developed by \citeauthor{ilyas2014microfilters}~\cite{ilyas2014microfilters} for classifying images present in tweets. Tweedr~\cite{ashktorab2014tweedr} is also a Twitter data mining tool that engages three operations in its data analysis pipeline: classification for classifying disaster specific tweets into categories such as damages and casualties, clustering for merging similar tweets and extraction for extracting words and phrases corresponding to damages and casualties.

\citeauthor{sakaki2012tweet}~\cite{sakaki2012tweet} studied the real-time nature of tweets for event detection. The authors proposed an earthquake reporting system that analyzes disaster related tweets for detecting and estimating the location of earthquake events (spatial estimation). They built a probabilistic spatio-temporal model using features such as the essential keywords in a tweet, the number of words in a tweet, and the context of the words specific to an event.

\subsubsection{Assessment of disasters}
\citeauthor{resch2018combining}~\cite{resch2018combining} presented an approach to combine semantic information extracted from tweets with spatial and temporal analysis to assess the affected areas and damages caused by natural disasters. The authors employed LDA for topic-modelling in two iterations---LDA was further applied to the results from the first iteration to identify granular topics. When validated against a manually annotated sample of tweets, their results showed that the affected areas and damages caused by disasters could be reliably identified by discovering similarities in spatial, temporal, and semantic details extracted from tweets.

Hierarchical multiscale analysis was conducted by \citeauthor{wu2018disaster}~\cite{wu2018disaster} to understand the role of socially generated data before, during and after a disaster and investigate if the joint use of such data alongside spatial information helps in disaster assessment. The authors used an opinion lexicon resource to compute sentiment polarity and performed hashtag and keyword frequency analysis for topic discovery. Results from the textual analysis were combined with spatio-temporal data to assess the damages done by Hurricane Sandy. The results showed the intensity of disaster-specific conversations originating from a region significantly correlating with the severity of damages.

\citeauthor{neppalli2017sentiment}~\cite{neppalli2017sentiment} showed that the sentiment of a population changes based on its distance from a disaster. The authors used lexicon methods to compute the polarity of tweets. Next, they classified polar tweets as positive and negative using NB and SVM classifiers. The study concluded that the mapping of sentiments during a disaster could uncover more vital situation awareness. Similarly, \citeauthor{kryvasheyeu2016rapid}~\cite{kryvasheyeu2016rapid} showed that the spatio-temporal analysis of Twitter data assists in the rapid assessment of disaster damage. The authors performed sentiment analysis on tweets created before, during and after Hurricane Sandy to examine if sentiment can be used as a predictor for assessing damages. The study showed negative sentiments correlating with damages and concluded that the proximity of the hurricane's path was highly correlated with tweet activity.

In~\cite{zhai2020examine}, \citeauthor{zhai2020examine} categorized neighbourhoods based on poor and non-poor attributes and used geo-tagged tweets alongside socio-demographic data to understand disaster situation awareness from the perspective of neighbourhood equity. They implemented LDA to identify specific topics associated with each neighbourhood and trained an LR model for sentiment analysis. Their results showed that poor areas are more likely to share negative opinions, as people from those communities are more likely to have their work affected during disasters.

\citeauthor{hernandez2019using}~\cite{hernandez2019using} used word vectors derived from a corpus of tweets to train a Bidirectional Long Short-Term Memory (biLSTM) network. They used a Conditional Random Field (CRF) output layer for improving classification accuracy. A toponym was formed based on labelled words and was geocoded and finally scored by a KDE function. The authors visualized the results on a map to identify concentration of disasters.

\subsubsection{Benchmarking}
Disaster specific tweets were analyzed by \citeauthor{ragini2018big}~\cite{ragini2018big} to identify the best feature set for sentiment analysis using: (i) BoW + Parts of Speech (POS) tagging and (ii) bigram and trigram subjective phrases. The authors used multiple lexicon-based methods to extract subjective tweets, as such tweets contain the sentiments of people compared to objective ones. Their results showed that the combination of subjective phrases and ML models trained on bigram features performs best for disaster specific Twitter data.

The performance of Bayesian network classifiers was benchmarked by \citeauthor{ruz2020sentiment}~\cite{ruz2020sentiment} against sentiment analysis on disaster specific tweets. The authors trained multiple learning models on tweets related to the Chilean earthquake and the Catalan independence referendum. Their results favoured SVM and RF models; however the tree augmented NB model was seen to produce competitive results, thus validating the applicability of Bayesian networks for analyzing sentiments of disaster-related tweets.

CNN models have been shown to outperform BoW models in the task of identifying informative messages from a stream of tweets during disastrous events. \citeauthor{caragea2016identifying}~\cite{caragea2016identifying} used the BoW model as a feature representation for training multiple SVM and ANN models using various combinations of $n$-grams. The results showed their CNN model outperforming all ANN and SVM models. The ANN model trained on unigrams and bigrams achieved the second-best accuracy; however, the margin was <2\%. Similar results were obtained in another study~\cite{nguyen2017robust}, where CNN models outperformed non-neural network models.

\begin{figure}[t]
  \centering
  \includegraphics[width=0.75\linewidth]{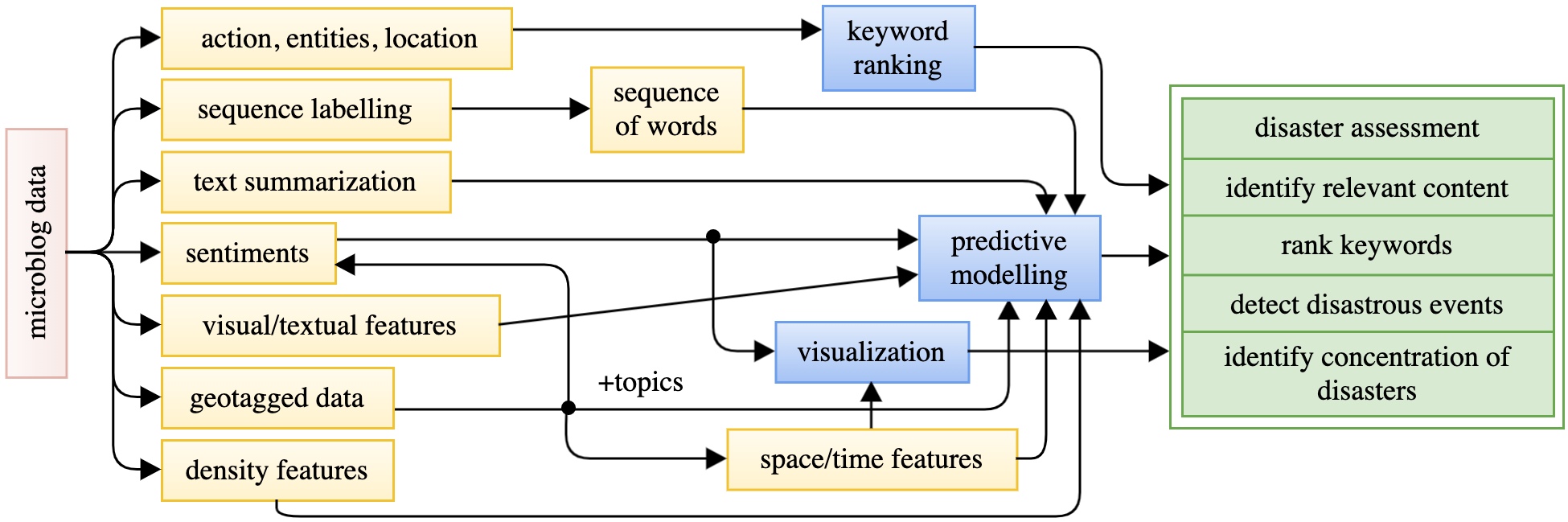}
  \caption{A methodological view of the literature in ``Disasters'' thematic area}
  \label{disasters-mindmap}
\end{figure}

An illustrative summary of methodologies discussed in this thematic area is given in Figure~\ref{disasters-mindmap}.

\subsection{Finance}
The stock markets, startups, and company-specific events are often discussed topics across social media. The discourse generated by financial communities on microblogs has been shown to correlate with stock options pricing, market movements, events' popularity, stock returns, and sustainable startups and business models. Table~\ref{lit-finances} provides an overview of the literature in this theme.

\begin{scriptsize}
\begin{table}
  \caption{Overview of the literature in ``Finance'' thematic area}
  \label{lit-finances}
  \begin{tabular}{p{0.9cm} p{0.3cm} p{0.3cm} p{2.4cm} p{1.65cm} p{1.2cm} p{4.6cm}}
    \toprule
    \textbf{Direction} & \textbf{Year} & \textbf{Study} & \textbf{Primary Dataset} & \textbf{Geo Scope} &  \textbf{Tasks} & \textbf{Best outcome}\\
    \midrule

\multirow{16}{1cm}{Stock market analysis} & 2011 & \cite{bollen2011twitter} & 9.85M tweets	& United States & \textit{Reg}, \textit{SnAn} & Direction Accuracy: 86.7\%\\

& 2012  & \cite{zhang2012predicting} &  3.8M retweets	&  United States & \textit{Reg}, \textit{SnAn} &	Retweets are predictive of market movements\\

& 2015 & \cite{liu2015social} & Twitter metrics of firms & United States &	\textit{AdSt, Clu} & Twitter metrics predict stocks comovement\\

 & 2016 & \cite{corea2016can} & $\approx$160k tweets & Global &	\textit{Reg}, \textit{SnAn} &	Tweets' volume is an essential feature for financial forecasting models. \\

& 2016 & \cite{zhang2016daily} & 50M tweets & Global & \textit{Reg}, \textit{SnAn} &  Happiness sentiments can Granger-cause the changes in index return\\

 & 2016 & \cite{wei2016twitter} & 3.28k Twitter volume
spikes & United States & \textit{AdSt} &	Volume spikes correlate with stock pricing when price change is extreme\\

 & 2017 & \cite{lopez2017impact} & 133k tweets & United States &	\textit{Reg}, SnAn &	Goodness-of-fit:  0.9776\\

& 2017 & \cite{you2017twitter} & Daily happiness index & Global &	\textit{Reg}, \textit{SnAn} &	Investors' sentiment correlate to stock returns\\

& 2018 & \cite{nisar2018twitter} & 60k tweets & United Kingdom &	\textit{Reg}, \textit{SnAn} &	Presence of correlation between public sentiment and investment behavior\\

\midrule

\multirow{6}{1.2cm}{Company-specific analysis} & 2017 & \cite{daniel2017company} & 192k tweets & United States & \textit{Cf},	\textit{SnAn} &	Case study specific results\\

& 2017 & \cite{leitch2017twitter} & 17k tweets & United States &	\textit{Reg}, \textit{SnAn} &	Negative relationship between sentiment and stock return.\\

& 2019 & \cite{singh2019analyzing} & 53k tweets & India &	\textit{TpMd}, \textit{SnAn} & Indian startup ecosystem is more inclined towards positive sentiments.\\

& 2019 & \cite{saura2019detecting} & 35k tweets & Global &	\textit{TpMd}, \textit{SnAn}	& Qualitative results presented\\

  \bottomrule
\end{tabular}
\end{table}
\end{scriptsize}

\subsubsection{Stock market analysis}
\citeauthor{liu2015social}~\cite{liu2015social} proposed a model for predicting stock comovement based on social media metrics. The authors reported that firms with official Twitter accounts had higher comovement. After analyzing data from the NYSE and the NASDAQ stock exchanges, they concluded that Twitter metrics, such as the number of followers and number of tweets created, highly correlate with the comovement of stocks. Further, they created homogeneous groups of firms using $K$-Means based on their Twitter metrics to show that they contribute to a better prediction of comovement than industry-specific label.

\citeauthor{wei2016twitter}~\cite{wei2016twitter} investigated the relationship between Twitter content volume spikes and stock pricing. They considered a volume spike when the number of tweets in a day was greater than the average number of tweets in the last $N$ days. They used the relative value of the number of tweets and added thresholds against the number of unique users and their diversity to avoid false volume spikes. Their analysis revealed that the volume spikes correlate with stock pricing when the change in the price is extreme.

\citeauthor{bollen2011twitter}~\cite{bollen2011twitter} extracted seven public mood time series from tweets to investigate if those mood descriptors are predictive of future stock values. The authors used Granger causality analysis to correlate the extracted mood time series with DJIA values. They reported that not all but some amongst the seven mood descriptors were Granger causative of the stock exchange's values. Similarly, \citeauthor{zhang2012predicting}~\cite{zhang2012predicting} tried to identify the relationship between Twitter data and financial market movement, such as gold price and stock market indicators. Six different opinion time series were extracted from a large-scale retweets dataset containing the following six keywords: \texttt{``dollar''}, \texttt{``gold''}, \texttt{``\$''}, \texttt{``job''}, \texttt{``economy''} and \texttt{``oil''}. The authors applied Granger causality analysis to the opinion time series versus financial market movement to show that the retweet information is correlated to and predictive of market movement.

\citeauthor{corea2016can}~\cite{corea2016can} analyzed Twitter data, concerning Apple, Facebook, and Google, to study if tweets can be considered representative of investors' sentiments. The author implemented Stepwise Regression for finding the best set of variables. The results showed that how much the public talks about a stock is more valuable than what they think about it. The authors observed the volume of tweets had a positive correlation with the stock price. In a similar study by \citeauthor{you2017twitter}~\cite{you2017twitter}, investors' sentiment showed to be a significant driving force on excess stock returns. Similarly, \citeauthor{lopez2017impact}~\cite{lopez2017impact} analyzed Twitter data to understand the influence of technical and non-technical investors on the stock market. The authors designed logit and probit models using the predictors derived from Twitter profiles such as experience in investing, number of followers, holding period. The study showed that the opinions of non-technical investors had relevance with the market risk, while for the technical ones, their opinions were insignificant.

\citeauthor{zhang2016daily}~\cite{zhang2016daily} extracted happiness indices from Twitter data for 11 international stock markets and compared it with index return. The results from their correlation regression model showed a positive influence of happiness sentiment on index return and opening/closing price of a trading session. Similarly, \citeauthor{nisar2018twitter}~\cite{nisar2018twitter} presented evidence of a correlation between public opinion and stock market price. The authors collected political-related tweets created before, during, and after the 2016 UK local elections and performed correlation and regression analysis on the sentiment of the tweets and the daily change in the price of FTSE 100. Their study showed promising results to support Twitter data as a medium for forecasting market price. 

\begin{figure}[t]
  \centering
  \includegraphics[width=0.75\linewidth]{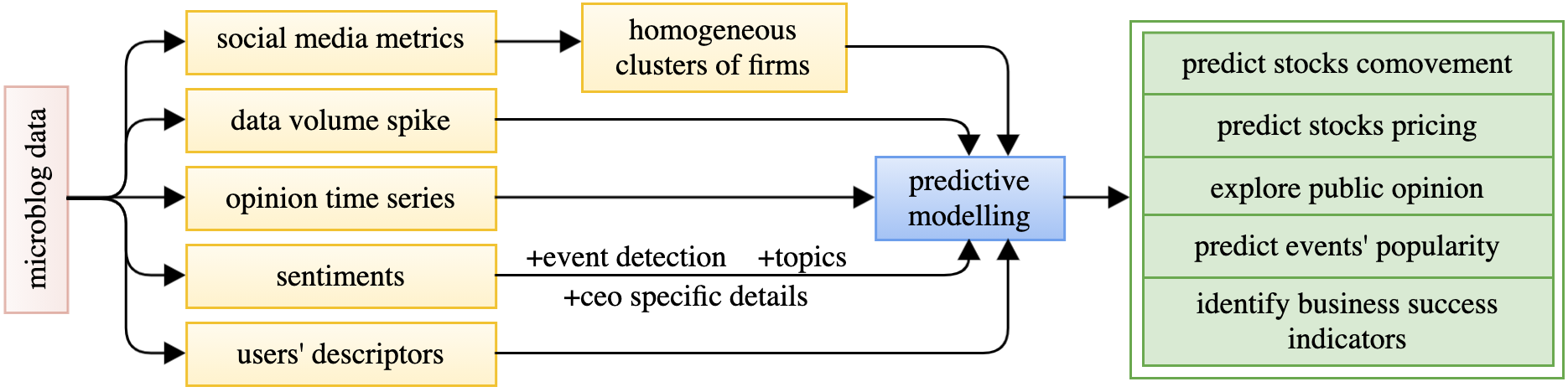}
  \caption{A methodological view of the literature in ``Finance'' thematic area}
  \label{finance-mindmap}
\end{figure}

\subsubsection{Company-specific analysis}
\citeauthor{daniel2017company}~\cite{daniel2017company} proposed a system to find event popularity through the sentiment of tweets concerning a company. Multiple public and a custom lexicon-based sentiment analyzers were used to compute sentiment scores. Event detection was done based on the peaks and drops observed in the average sentiment of the tweets. One of their case studies involving a Fortune 500 company showed people's excitement during the launch of product A relatively higher than product B, where B was a minor upgrade to A.

\citeauthor{leitch2017twitter}~\cite{leitch2017twitter} designed regression models based on predictors such as public sentiment during the announcement of a new Chief Executive Officer (CEO), location of the company, number of employees and various features associated with the CEO, including gender, degree details, and experience, to study if Twitter sentiment yields any relationship with stock returns. Their results reported that the sentiment score had a significant negative correlation with stock returns, while the age of the CEO had a significant positive correlation. Features such as gender, experience and number of employees reported positive correlation but were statistically insignificant.

\citeauthor{singh2019analyzing}~\cite{singh2019analyzing} used Twitter data to analyze the startup ecosystem in India. The authors collected tweets from 15 different startups across multiple industries and applied NB for sentiment analysis and LDA for topic modelling. They reported that the Indian startup ecosystem was inclined towards positive sentiments as the startups were more concerned about digital technologies, people, planet and profit. More specifically, startups in cities with the availability of resources, connectivity and a huge market base seemed to be finding it easy to do business. Similarly, \citeauthor{saura2019detecting}~\cite{saura2019detecting} used SVM for sentiment analysis and LDA for topic modelling on tweets using hashtags related to startups to identify indicators for a successful startup business. Their study showed that the founders' attitudes and methodologies were associated with positive sentiments. In contrast, frameworks, and programming languages were some of the factors inducing negative sentiments.

An illustrative summary of methodologies discussed in this thematic area is given in Figure~\ref{finance-mindmap}.

\subsection{Physical Environment}

While considering microblog users as ``environment monitoring sensors'', researchers have applied a wide range of analytical strategies that exploit microblog data to generate valuable insights concerning traffic patterns, conditions, and incidents, weather situations, outdoor air pollution, climate change, and global warming. Table~\ref{lit-environment} provides an overview of the literature in this theme.

\subsubsection{Weather and Climate}

In ~\cite{giuffrida2020assessing}, \citeauthor{giuffrida2020assessing} analyzed tweets containing weather-related keywords alongside meteorological data to show that Twitter can be regarded as an alternative data source for assessing the effect of weather on human outdoor perception. The authors trained an opinion-based classifier to classify tweets as relevant or irrelevant and further categorize the relevant ones into neutral, positive or negative. Tweets were next analyzed together with meteorological data to establish comfort ranges. The ranges identified through tweets analysis were found to be in good agreement with the ranges derived through questionnaires and interviews.

\citeauthor{kirilenko2014public}~\cite{kirilenko2014public} reported the US, UK, Canada, and Australia to have created more discourse related to climate change on Twitter than the rest of the world. The study performed spatio-temporal and network analyses to identify daily patterns of discourse, the significant events affecting the discourse, and the most influential media houses and Twitter users. In~\cite{jang2015polarized}, \citeauthor{jang2015polarized} performed a content analysis on tweets related to climate change originating from the US, UK, Canada, and Australia between 2012--14. The authors constructed five representative frames, namely Real, Hoax, Cause, Impact, and Action, based on associated keywords to extract the relative prevalence of each frame, country-wise. The study summarised that hoaxes are common, and the Cause, Impact and Action-based Twitter discourse are comparatively lower in the US.

\citeauthor{chen2019detecting}~\cite{chen2019detecting} analyzed similar climate change-related tweets to explore the temporal details of climate change discourse and identify Twitter users that deny the existence of climate change. The related English keywords were translated into 34 different languages, and the resulting set of keywords were used for filtering the tweets. A sample of 2k tweets was annotated to train an ANN model for classifying the users as deniers and non-deniers. Their exploratory analysis showed that the discussion about climate change is driven by extreme weather events and changes in policies. 

\subsubsection{Pollution}

In~\cite{hswen2019feasibility}, \citeauthor{hswen2019feasibility} studied if Twitter data can be used for monitoring outdoor air pollution. The authors collected geo-tagged tweets (non-media) containing air pollution terms from Greater London and performed sentiment analysis on the tweets using VADER\footnote{https://github.com/cjhutto/vaderSentiment
}. Next, cross-correlation analysis was done to find relationships between sentiment trends and levels of PM2.5 (particles with diameter less than 2.5 micrometres). Their results showed a significant correlation between negative sentiment tweets and PM2.5 data, thereby hinting that Twitter users from a densely populated area can be treated as ``social sensors'' of PM2.5 levels.

\citeauthor{sachdeva2018using}~\cite{sachdeva2018using} studied the effectiveness of using Twitter data in ascertaining the impacts of wildfire events on air quality. The authors used structural topic model for topic modelling. Topics related to ``smoke'' were then identified for tagging tweets. Their spatio-temporal model used this tagged information for assessing the relationship between the frequency of smoke-related tweets with daily PM2.5 levels. Their results showed that the tweets relating to smoke were better predictors of air quality than tweets related to more generic wildfire discourse. 

\begin{scriptsize}
\begin{table}[t]
  \caption{Overview of the literature in ``Physical Environment'' thematic area}
  \label{lit-environment}
   \begin{tabular}{p{1.2cm} l l p{2.5cm} p{2.4cm} p{1.7cm} p{2.5cm}}
    \toprule
    \textbf{Direction} & \textbf{Year} & \textbf{Study} & \textbf{Primary Dataset} & \textbf{Geo Scope} & \textbf{Tasks} & \textbf{Best outcome}\\
    \midrule

\multirow{4}{1cm}{Weather and climate analysis} & 2014 & \cite{kirilenko2014public} & 1.85M tweets & Global &	\textit{SpTmMd}, \textit{NetAn} &	Qual. disc. presented\\

& 2015 & \cite{jang2015polarized} & Rnd smp. of 500 tweets & Global & \textit{CoFqAn}	& Precision: 96\%\\

& 2019 & \cite{chen2019detecting} & 2k tweets & Global	 & \textit{WoVec}, \textit{Cf} &	Accuracy:  88\%\\

& 2020 & \cite{giuffrida2020assessing} & 38k tweets	& United States & \textit{SnAn} & Class. Accuracy:  88\%\\

\midrule

\multirow{2}{1cm}{Pollution analysis} & 2018 &  \cite{sachdeva2018using} & 39k geo-tagged tweets & California &	\textit{TpMd} &	Stat. analysis presented\\

& 2019 & \cite{hswen2019feasibility} & 60k tweets & Greater London &	\textit{SnAn}, \textit{AdSt} &	Corr. coef.: 0.816\\

\midrule

\multirow{10}{1cm}{Road and traffic analysis} & 2010 & \cite{carvalho2010real} &	565k tweets & Portugal &	\textit{Cf} &	F-measure: 0.96\%\\

& 2015 & \cite{wang2015citywide} & 245k tweets & Chicago &	\textit{SpTmMd}, \textit{AdSt} &	Stat. analysis presented\\

& 2015 & \cite{gong2015identification} & System self-collects data & Melbourne &	\textit{Clu} &	Qual. disc. presented\\

& 2015 & \cite{tejaswin2015tweeting} & 200 tweets & India &	\textit{NER} &	F-measure: 0.918\\

& 2015 & \cite{d2015real} &	1.33k/999 tweets	& Italy & \textit{Cf} &	F-measure: 0.957\\

& 2016 & \cite{gu2016twitter} &	10k/11k tweets	&  Pittsburgh, Philadelphia & \textit{Cf} &	F-measure: 0.95\\

& 2017 & \cite{wang2017real} & 9.70k tweets & United Kingdom &	\textit{TpMd}, \textit{Cf} &	Recall: >90\%\\

& 2019 & \cite{dabiri2019developing} & 51k tweets	& United States & \textit{WoVec}, \textit{Cf} &	F-measure: 0.986\\

& 2020 & \cite{essien2020deep} & 9k geo-tagged tweets & United Kingdom	& \textit{Reg} &	MAE: 5.5\\

& 2021 & \cite{yao2021twitter} & 1.78M  tweets	 & Pittsburgh &  \textit{SnAn}, \textit{Clu} &	Accuracy: $0.88\pm0.05$\\

\bottomrule
\end{tabular}\\
\end{table}
\end{scriptsize}

\subsubsection{Traffic}

\citeauthor{tejaswin2015tweeting}~\cite{tejaswin2015tweeting} designed an incident alert and mapping system by extracting location entities from tweets that describe traffic situations. To deal with the sparsity of data, the authors created multiple grids of the overall city area and grouped traffic incidents to generate statistics on historical data for extracting real-time insights. Predictors such as weather data, grid number and temporal detail were used for training an RF classifier for predicting traffic incidents.

\citeauthor{essien2020deep}~\cite{essien2020deep} mined geo-tagged Twitter data together with traffic and weather information to study if such data can improve urban traffic flow prediction. The authors trained a biLSTM stacked autoencoder model and evaluated it on an urban road network in the UK. They prioritized the tweets created from the official Twitter accounts of road-traffic organizations to deal with authenticity and veracity issues. Their results showed that combining Twitter data with traffic, rainfall and temperature data assists in a design of a more accurate traffic flow prediction model.

\citeauthor{wang2015citywide}~\cite{wang2015citywide} proposed a traffic congestion estimation framework by using information regarding traffic from Twitter. A frequent-pattern spatial and temporal analysis was done to identify the road networks that are more likely to have co-occurring congestions. Their estimation approach also considered social events within the city as supplementary information. The authors used the time and location of such events to model the impact on traffic in the nearby roads using a Gaussian distribution. Similarly, \citeauthor{yao2021twitter}~\cite{yao2021twitter} extracted sleep-wake status, local event details and traffic incidents as tweeting profiles and used the following generalizations to build a prediction framework for next morning commute congestion: (i) roads are congested the following morning if people sleep earlier, (ii) lower travel demand the following day if the earlier evening had social events. Their approach outperformed existing approaches that excluded Twitter data. In a similar study, \citeauthor{gong2015identification}~\cite{gong2015identification} utilized created time and geocoordinates information extracted from tweet data dictionaries to identify a road as congested if multiple tweets share similar spatial and temporal attributes. The authors used a density-based clustering and assumed the presence of congestion if 4 or more tweets are created in 1 km stretch of road within a 15-minute time window.

\citeauthor{dabiri2019developing}~\cite{dabiri2019developing} proposed a traffic information system that processes a large volume of Twitter data and detects traffic events. The authors used word2vec and FastText for feature representation to train three deep learning architectures (CNN, LSTM and CNN+LSTM) to classify traffic-related information into three categories: traffic incident, traffic information and condition, and irrelevant. Their results showed the CNN model trained on word2vec outperforming all other models, including the SVM, RF, and MLP models proposed by \citeauthor{carvalho2010real}~\cite{carvalho2010real}, the NB, DT, and SVM models proposed by \citeauthor{d2015real}~\cite{d2015real}, and the semi-NB model proposed by \citeauthor{gu2016twitter}~\cite{gu2016twitter}.

Identifying relevant tweets amongst those created by official accounts of road administration offices and that mention road/street is simple compared to tweets created informally from drives, as such tweets are likely to have a significant amount of noise. To deal with this problem, \citeauthor{wang2017real}~\cite{wang2017real} first filtered the tweets based on a set of automatically generated keywords and then applied LDA to achieve higher recall scores. Binormal separation and NB word weighting schemes were used to identify the top $N$ set of keywords.

\begin{figure}[t]
  \centering
  \includegraphics[width=0.75\linewidth]{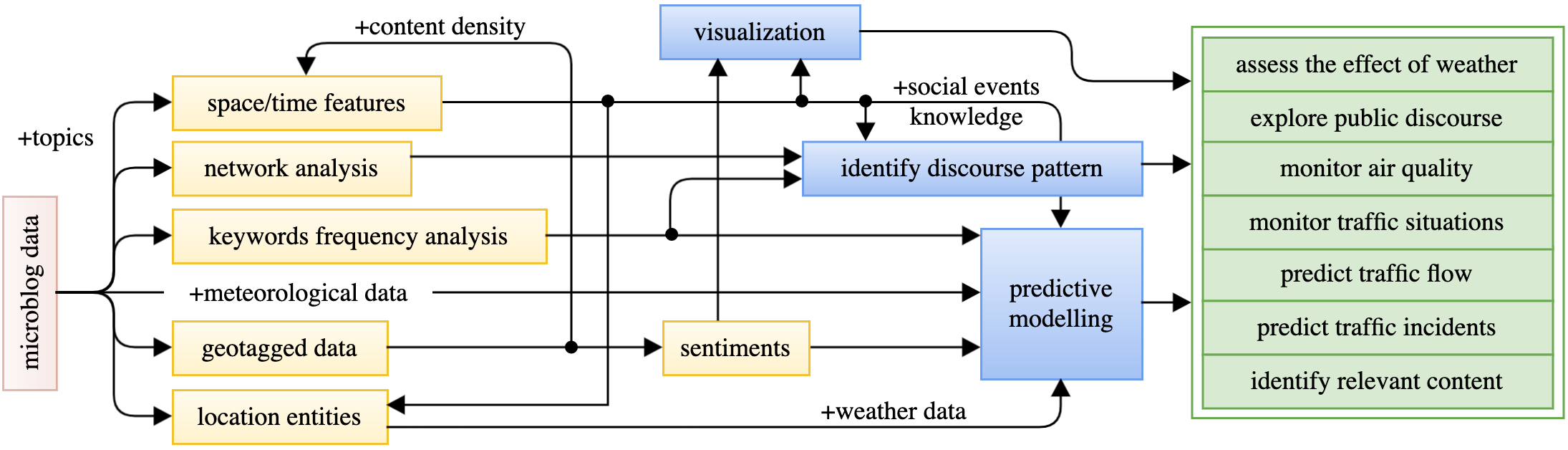}
  \caption{A methodological view of the literature in ``Physical Environment'' thematic area}
  \label{environment-mindmap}
\end{figure}

An illustrative summary of methodologies discussed in this thematic area is given in Figure~\ref{environment-mindmap}.

\subsection{Politics}
Researchers have analyzed microblog data to study the popularity of election candidates, predict election outcomes, understand voting intentions of a nation, detect emerging political topics, identify probable riot events, and understand human/bot communication patterns during disruptive events. Table~\ref{lit-politics} provides an overview of the literature in this theme.

\subsubsection{Election Analysis}

Twitter has been reported as an essential source of data for complementing traditional polls and surveys for political forecasting. In~\cite{tumasjan2011election}, \citeauthor{tumasjan2011election} performed sentiment analysis and generated multidimensional profiles of politicians using tweets related to the 2009 German Federal Election. Their results showed the tweets' sentiments corresponding closely with voters' political inclinations and the relative volume of party-specific tweets as a strong predictor of the federal election results. \citeauthor{bermingham2011using}~\cite{bermingham2011using} also showed the volume of tweets as a significant predictive variable in predicting election outcomes. In~\cite{gaurav2013leveraging}, \citeauthor{gaurav2013leveraging} reported that the volume-based approach in election prediction could be improved by also considering different aliases used for a candidate while constructing datasets for analysis.

\citeauthor{wang2012system}~\cite{wang2012system} designed a real-time tweets sentiment analysis system targeting the 2012 US Presidential Election. The authors used around 200 rules for filtering tweets related to the election campaign. A sample of relevant tweets was crowd-sourced for polarity labelling to train an NB classifier. Sentiment trend and volume of tweets associated with each candidate were aggregated for visualization on their dashboard. Similar sentiment-based approaches have also been used in forecasting the winners of the 2013 Pakistani Election, the 2014 Indian Election~\cite{kagan2015using}, the 2015 UK General Election~\cite{burnap2016140}, the 2016 Indian General State Elections~\cite{sharma2016prediction}, and the 2017 French Presidential Election~\cite{wang2017prediction}.

In~\cite{yaqub2017analysis}, \citeauthor{yaqub2017analysis} analyzed the 2016 US presidential election Twitter discourse to evaluate how well such data represent public opinion. The keywords \texttt{``trump''}, \texttt{``clinton''}, and \texttt{``election2016''} were used for pulling relevant tweets. The authors reported negative trends for both the candidates and quoted reports that termed the 2016 election as ``the most negative campaign'' to support their results. Their subsequent analysis showed that tweets containing only the keyword \texttt{``trump''} had a lower average negative score than the tweets associated to \texttt{``clinton''}. A similar methodology was used in an earlier study by \citeauthor{ibrahim2015buzzer}~\cite{ibrahim2015buzzer}, where only the positive sentiment tweets were considered candidate-wise for predicting the winner of the 2014 Indonesian Presidential Election.

\citeauthor{song2014analyzing}~\cite{song2014analyzing} analysed tweets related to the 2012 Korean Presidential Election to show Twitter as an essential medium for tracking topical trends. The authors performed temporal LDA for topic modelling and a term co-occurrence analysis for tracing chronologically co-occurring topics. They also implemented network analysis based on ``user-mentions'' to identify thematic connections among users. \citeauthor{rill2014politwi}~\cite{rill2014politwi} designed a system for early detection of current and emerging political topics using Twitter. The authors considered Twitter hashtags as candidates for emerging topics by considering the temporal change in the number of tweets associated with each hashtag, i.e. comparing the current number of tweets $N(H,t)$ associated with a hashtag $H$ to the number of tweets in the previous period $N_{p}(H,t-1)$. Their results showed that emerging topics on Twitter could be detected immediately after their occurrence, and these topics seemed to appear earlier on Twitter compared to Google Trends. 

\citeauthor{paul2017compass}~\cite{paul2017compass} designed SVM and LR classifiers for identifying political and non-political tweets and performed sentiment analysis on the political ones concerning their political alliances (Democratic or Republican). The authors used LDA and word2vec for extracting enriched keywords specific to politics and political alliances for labelling tweets. They trained unigram and unigram-bigram based models using SVM and LR. Sentiment Analysis was performed on tweets at the political affiliation level, for which the authors showed the FastText based model outperforming SVM, LR, NB and LSTM models.

\citeauthor{barbera2015birds}~\cite{barbera2015birds} used a Markov Chain Monte Carlo (MCMC) based method to estimate the political ideology of people by exploiting their ``following'' profiles on Twitter. The author used an NoU-Turn sampler and a random-walk Metropolis-Hastings algorithm for estimating the parameters $\alpha_{j}$ (measure of popularity of $j$) and $\beta_{i}$ (measure of political interest of $i$) indexed by $j$ and $i$ respectively, where $i$ and $j$ relate to a sample of 10k $i$-users that follow at least 10 $j$-users. The proposed method distinguishes the political orientation of both political figures and ordinary citizens with their location on the ideological scale associated with ``left-wing'' and ``right-wing'' politics.

\begin{scriptsize}
\begin{table}[t]
  \caption{Overview of the literature in ``Politics'' thematic area}
  \label{lit-politics}
  \begin{tabular}{p{1.2cm} l l l p{2.4cm} l p{2.6cm}}
    \toprule
    \textbf{Direction} & \textbf{Year} & \textbf{Study} & \textbf{Primary Dataset} & \textbf{Geo Scope} & \textbf{Tasks} & \textbf{Best outcome}\\
    \midrule

\multirow{14}{1cm}{Election Analysis} & 2011 & \cite{tumasjan2011election}  & 100k tweets & Germany &	\textit{SnAn}, \textit{CoFqAn} &	MAE: 1.65\% \\

 & 2011 & \cite{bermingham2011using} & 7.2k tweets & Ireland &	\textit{SnAn}, \textit{CoFqAn}, \textit{Reg} &	MAE: 3.67--5.85\% \\

 & 2012 & \cite{wang2012system} & 17k tweets & United States &	\textit{SnAn}, \textit{Cf} &	Accuracy: 59\%\\

& 2013 & \cite{gaurav2013leveraging} & 62k tweets & Latin America &	\textit{CoFqAn} &	Average RMS: <0.03\\

& 2014 & \cite{song2014analyzing}  & 1.73M tweets & South Korea &	\textit{TpMd}, \textit{NetAn} &	-\\

& 2014 & \cite{rill2014politwi} & 4M tweets & Germany &	\textit{CoFqAn}, \textit{SnAn} &	Corr. coef.: 0.68\\

& 2015 & \cite{barbera2015birds}  & 759k Twitter users & United States &	\textit{AdSt} &	Stat. analysis presented\\

& 2015 & \cite{ibrahim2015buzzer} & 10M tweets &	Indonesia & \textit{SnAn} &	MAE: 0.61\%\\

& 2015 & \cite{kagan2015using} & 23M tweets & India, Pakistan &	\textit{SnAn} &	Corr. coef.:  0.83--0.986\\

& 2016 & \cite{sharma2016prediction} & 42k tweets & India &	\textit{SnAn} &	Accuracy: 78.4\% \\

& 2016 & \cite{burnap2016140} & 13.8M & United Kingdom &	\textit{SnAn} &	Stat. analysis presented\\

& 2017 & \cite{yaqub2017analysis} &  3.1M tweets & United States &	\textit{SnAn} &	-\\

& 2017 &  \cite{paul2017compass}  & 2M geo
tweets & United States &	\textit{TpMd}, \textit{SnAn}, \textit{SpTmMd} &	F-measure: 0.930\\

& 2017 & \cite{wang2017prediction} & Not-specified & France &	\textit{SnAn} &	2\% difference from fact\\

\midrule
\multirow{3}{1cm}{Riots Analysis} & 2017 & \cite{alsaedi2017can} & 40M/1.6M tweets & England &	\textit{NER}, \textit{SnAn}, \textit{Cf}, \textit{Clu} &	F-measure: 85.43\%\\

& 2020 & \cite{kuvsen2020you} & 4.5M tweets & Germany, United States, Catalonia &	\textit{SnAn}, \textit{NetAn} &	Stat. analysis presented\\

\bottomrule
\end{tabular}\\
\end{table}
\end{scriptsize}

\subsubsection{Riots Analysis}

\citeauthor{alsaedi2017can}~\cite{alsaedi2017can} proposed an event detection framework for detecting probable disruptive events such as riots using Twitter. The framework utilizes an NB classifier to filter event-related tweets from non-event tweets. An online clustering algorithm then groups the extracted event-related tweets to identify potentially disruptive events based on spatial, temporal and textual features extracted from the tweets. The textual features list also included cosine similarity, sentiment polarity, mention ratio, hashtag ratio, and URL ratio. Next, a temporal TF-IDF is used for summarizing and representing the topics discussed in each cluster.

\citeauthor{kuvsen2020you}~\cite{kuvsen2020you} performed lexicon-based sentiment analysis on tweets using manually selected keywords and hashtags to identify the influence of bots during riot events. The authors used scores computed by Botometer\footnote{https://botometer.osome.iu.edu/} to distinguish if an account is a bot. They further constructed two types of directed communication networks, a human-accounts-only network and an all-accounts network, using the ``mentions'' information available in the tweets. Their findings showed that bot accounts were more involved in amplifying negative emotions to influence the overall sentiment of the discourse. The bot accounts were also noticed receiving significant social interactions from human accounts as the bot accounts seemed to be regarded as credible sources of information.

\begin{figure}[t]
  \centering
  \includegraphics[width=0.75\linewidth]{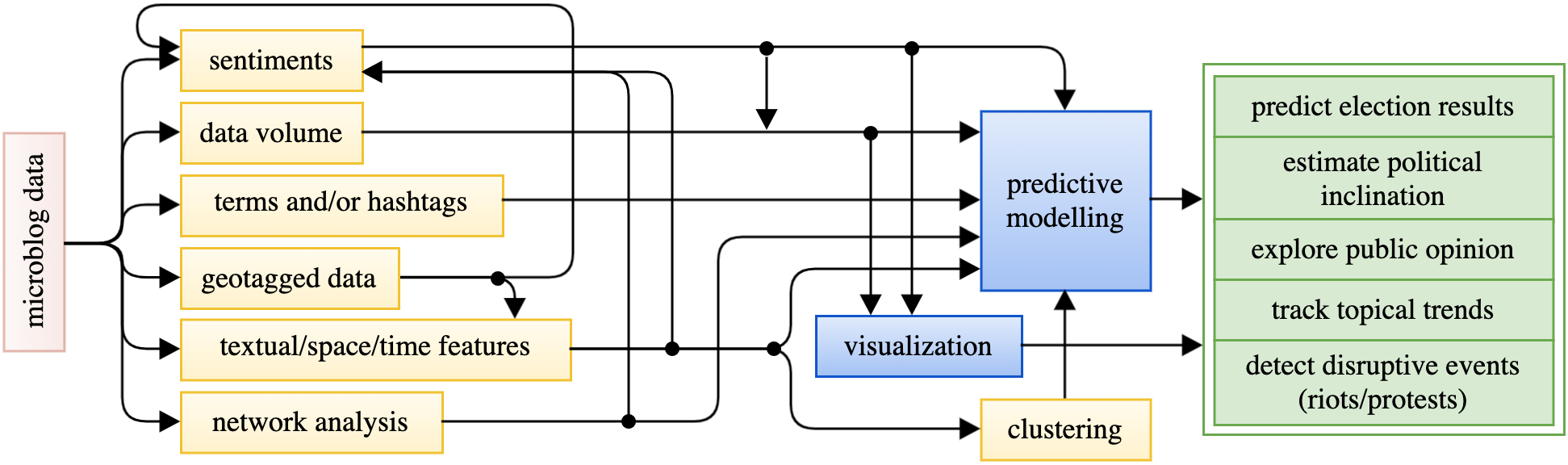}
  \caption{A methodological view of the literature in ``Politics'' thematic area}
  \label{politics-mindmap}
\end{figure}

An illustrative summary of methodologies discussed in this thematic area is given in Figure~\ref{politics-mindmap}.

\subsection{Health and Population}
Public health-related microblog data concerning epidemics such as Influenza, Zika, Dengue, prescription drug abuse, and the coronavirus disease (COVID-19) pandemic has been used for predicting new case counts, extracting detailed insights of the illness/abuse, tracking the flow of conspiracy theories and misinformation. Table~\ref{lit-health} provides an overview of the literature in this theme.

\subsubsection{Epidemic Analysis}
Based on flu related tweets collected between 2009 and 2010, \citeauthor{achrekar2011predicting}~\cite{achrekar2011predicting} designed auto-regression models to show that the volume of tweets highly correlate with the number of influenza-like illness cases. Also, \citeauthor{masri2019use}~\cite{masri2019use} used auto-regressive models on weekly Zika epidemic case counts and tweets for estimating the number of cases, on a weekly basis, one week in advance.

\citeauthor{aramaki2011twitter}~\cite{aramaki2011twitter} trained an SVM classifier to identify true influenza tweets and used them as a predictor for influenza epidemics detection. Their method outperformed approaches that considered the use of overall volume of tweets or used Google trends data. Similarly, \citeauthor{de2017dengue}~\cite{de2017dengue} tracked \texttt{``Dengue''}, \texttt{``aedes''} and \texttt{``aegypti''} keywords to collect tweets and employed a classifier to identify tweets related to personal experience with Dengue. The authors used the total number of captured tweets and Dengue cases to design a generalized additive model. Their results showed that tweets have a strong potential to be an explanatory variable for Dengue estimation models.

\citeauthor{mamidi2019identifying}~\cite{mamidi2019identifying} trained multiple word2vec and $n$-gram-based LR, SVM and RF models for analyzing sentiments of tweets concerning Zika and further identified the main topics within the negative sentiment tweets to distinguish the significant concerns of the targeted population.

Twitter data has been observed as useful for providing insights into the prescription drug epidemic~\cite{shutler2015drug}. \citeauthor{phan2017enabling}~\cite{phan2017enabling} collected tweets concerning well-known prescription and illegal drugs and trained SVM, DT, RF and NB classifiers to identify signals of drug abuse in the tweets. Their proposed drug abuse detection model could identify relationships between \texttt{``Abuse''} and \texttt{``Fentanyl''}, \texttt{``Addiction''} and \texttt{``Heroin''}, etc. Similarly, \citeauthor{katsuki2015establishing}~\cite{katsuki2015establishing} studied the relationship between Twitter data and promotion of non-medical use of prescription medications (NUPM). The authors trained an SVM classifer on a sample of labelled data for identifying NUPM-relevant tweets. Their results showed a significant proportion of tweets promoting NUPM that actively marketed illegal sales of prescription drugs of abuse. A similar result was reported by \citeauthor{sarker2016social}~\cite{sarker2016social}, where NB, SVM, Maximum Entropy and DT classifiers were trained for identifying tweets containing signals of medication abuse.

\begin{scriptsize}
\begin{table}[t]
  \caption{Overview of the literature in ``Health and Population'' thematic area}
  \label{lit-health}
 \begin{tabular}{p{1.2cm} l l p{2.6cm} p{2.1cm} p{1.9cm} p{2.5cm}}
    \toprule
    \textbf{Direction} & \textbf{Year} & \textbf{Study} & \textbf{Primary Dataset} & \textbf{Geo Scope} & \textbf{Tasks} & \textbf{Best outcome}\\
    \midrule

\multirow{9}{1cm}{Epidemic Analysis} & 2011 & \cite{achrekar2011predicting}  & 4.7M tweets & Global &	\textit{CoFqAn}, \textit{Reg} &	Corr. coef.: 0.9846\\

& 2011 & \cite{aramaki2011twitter} & 0.4M tweets &	Global & \textit{AdSt}, \textit{CoFqAn}, \textit{Cf} &	Corr. coef.: 0.89\\

& 2015 & \cite{katsuki2015establishing} & 2.4M tweets & United States &	\textit{Cf} &	Accuracy: 93.5--95.1\%\\

& 2016 & \cite{sarker2016social} & 129k tweets & Global &	\textit{Cf} &	Accuracy: 85\%\\

& 2017 & \cite{de2017dengue} & 1.65M tweets	& Brazil & \textit{CoFqAn}, \textit{Cf}, \textit{Reg} &	MRE: 0.345\\

& 2017 & \cite{phan2017enabling}  & 31k tweets & United States &	\textit{Cf} &	F-measure: 0.746\\

& 2019 & \cite{mamidi2019identifying} & 48k tweets & Zika affected regions &	\textit{Cf}, \textit{TpMd}, \textit{WoVec} &	F-measure: 0.68\\

& 2019 & \cite{masri2019use}  & Weekly basis tweets count & United States &	\textit{Reg}, \textit{SpTmMd} &	$R^{2}$: 0.74\\

\midrule
\multirow{9}{1cm}{Pandemic Analysis} & 2020 & \cite{chandrasekaran2020topics} & 13.9M tweets & Global &	\textit{TpMd}, \textit{SnAn}	& Stat. analysis presented\\

& 2020 & \cite{abd2020top} & 167k tweets & Global &	\textit{Cf}, \textit{SnAn}, \textit{TpMd} &	Stat. analysis presented\\

& 2020 & \cite{li2020data} & 115k Weibo posts & Wuhan &	\textit{CoFqAn}, \textit{Reg} &	$R^{2}$: 0.621\\

& 2020 & \cite{shen2020using}  & 14.9M Weibo posts & China &	\textit{Cf}, \textit{CoFqAn} &	F1-measure: 0.880\\

& 2020 & \cite{xue2020hidden} & 1M tweets & Global &	\textit{TpMd} &	Qual. disc. presented\\

& 2020 & \cite{ahmed2020covid} & 10k tweets, 6.55k users &	United Kingdom & \textit{CoFqAn}, \textit{NetAn} &	Stat. analysis presented\\

& 2021 & \cite{gerts2021thought}  & 1.8M tweets & Global &	\textit{Cf}, \textit{SnAn}, \textit{TpMd} &	F-measure: 0.857\\

& 2021 &  \cite{saleh2021understanding} & 574k tweets & Global &	\textit{SnAn} &	Stat. analysis presented\\

\bottomrule
\end{tabular}
\end{table}
\end{scriptsize}

\subsubsection{Pandemic Analysis}
\citeauthor{chandrasekaran2020topics}~\cite{chandrasekaran2020topics} used LDA for discovering topics and VADER for exploring associated sentiments of COVID-19 related tweets. They reported that the average sentiment for topics such as growth and spread of cases, symptoms, racism and political impact was negative throughout the analysis period. In contrast, the sentiment was seen shifting towards positivity for government response, impact on the health system, economy, and treatment and recovery. Similarly, \citeauthor{abd2020top}~\cite{abd2020top} performed topic-wise sentiment analysis using LDA and TextBlob\footnote{https://textblob.readthedocs.io/en/dev/} to show the presence of negative sentiment for deaths and racism-related topics.

\citeauthor{saleh2021understanding}~\cite{saleh2021understanding} collected tweets containing \texttt{\#socialdistancing} and \texttt{\#stayathome} hashtags for identifying tweets specific to social distancing. They used TextBlob for sentiment and subjectivity analysis. Based on the dominance of positive and subjective tweets, the authors concluded that Twitter users supported the social distance measures. \citeauthor{xue2020hidden}~\cite{xue2020hidden} collected family violence-related tweets by tracking keywords such as \texttt{``family violence''} and \texttt{``child abuse''}, and performed topic modelling using LDA. They identified 9 different topical themes, including violence types, risk factors, victims, law enforcement responses, social awareness, etc. The study showed topic modelling as an essential technique for monitoring violence related situations through tweets.

\citeauthor{gerts2021thought}~\cite{gerts2021thought} used an RF classifier for identifying tweets concerning COVID-19 conspiracy theories related to 5G technology, Bill \& Melinda Gates Foundation, origin of the virus, and COVID-19 vaccines. They showed that a significant amount (>40\% except for the ``vaccines'') of tweets were spreading misinformation. They used lexicon-based methods for sentiment analysis to report that negative sentiment is most prevalent in the tweets classified as misinformation. Similarly, \citeauthor{ahmed2020covid}~\cite{ahmed2020covid} collected tweets concerning the 5G and COVID-19 conspiracy by tracking the \texttt{\#5GCoronavirus} hashtag. The results from their network analysis hinted a lack of authorities in combating misinformation, and content analysis revealed a significant amount (34.8\%) of tweets containing views that 5G and COVID-19 were linked.

The volume of socially generated data has been observed predictive of the number of COVID-19 cases. In~\cite{li2020data}, \citeauthor{li2020data} analyzed Chinese language Weibo posts originating from Wuhan and built a linear regression model to show a positive correlation between the number of Weibo posts and the daily cases of COVID-19 in Wuhan. Similarly, \citeauthor{shen2020using}~\cite{shen2020using} designed an RF classifier for identifying Weibo posts that reported anything related to symptoms and diagnosis (``sick messages'') of COVID-19. The authors performed Granger causality test and observed that the number of ``sick messages'' could significantly predict the daily number of cases, 14 days in advance.

\begin{figure}[t]
  \centering
  \includegraphics[width=0.75\linewidth]{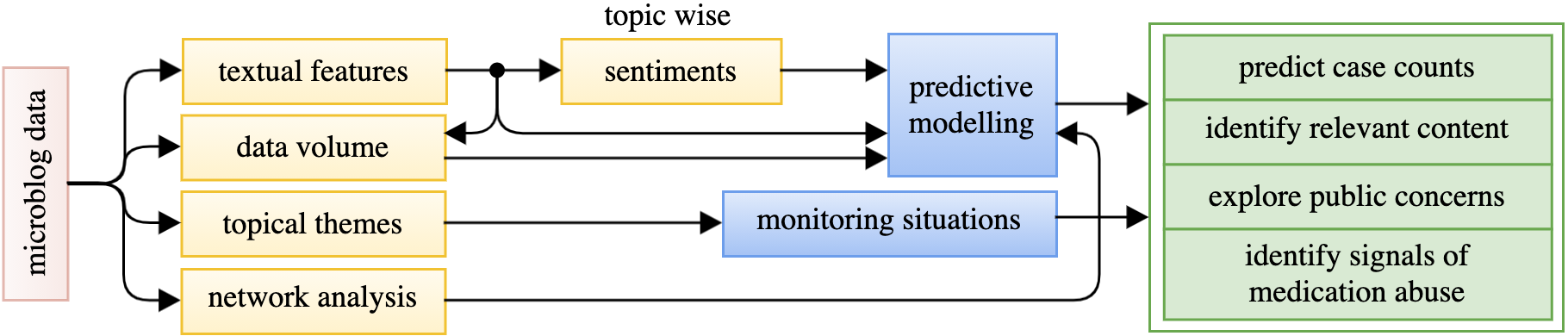}
  \caption{A methodological view of the literature in ``Health and Population'' thematic area}
  \label{health-mindmap}
\end{figure}

An illustrative summary of methodologies discussed in this thematic area is given in Figure~\ref{health-mindmap}.

\section{Revisiting the Literature}
\label{discussion}

\subsection{Data Corpus}
Microblogs provide multiple API endpoints for accessing the public feed generated on their platforms. For instance, Twitter allows two API types for accessing its content: a \textit{search API} for searching against a sample of tweets created in the past 7 days (standard version) \cite{lamsal2021design}, and a \textit{streaming API} for accessing its real-time feed. Similarly, Weibo provides developers with a \textit{Timeline API} for returning the latest Weibo posts and a \textit{search API}, which is restricted to a set of approved developers, for searching historical Weibo posts~\cite{weibo2021}. Furthermore, microblogs typically provide endpoints associated with followers, friends, geo location, favorites, reshare, trends, etc.

In general, the literature employs two methods for curating data corpus. The first method involves using official API endpoints to collect near real-time microblog content, or search for historical content. The second method involves hydrating content identifiers to gather complete information. Second method is employed if a microblogging platform only allows sharing of identifiers, such as Twitter \cite{lamsal2020coronavirus}, where researchers are restricted in sharing data other than tweet identifiers \cite{lamsal2021design}.

\subsection{The Deep Learning Trend}
\label{deep-trend}

During the survey, we observed that almost every study had its own set of data. Implementing the same methodology on different datasets generates mixed outcomes, so quantitatively comparing the outcomes of the surveyed papers is difficult to justify. However, the tasks implemented by the surveyed papers concentrate on the textual and multimedia context of microblog data; therefore, the advancements in NLP and Computer Vision undeniably expand the boundary of microblog analytics. The ongoing radical improvements in hardware have revolutionized the deep learning sphere. Deep models seem to consistently achieve state-of-the-art outcomes in various tasks, including text classification \cite{yang2019xlnet}, sentiment analysis \cite{raffel2019exploring, jiang2019smart}, short text clustering \cite{zhang2021supporting}, image classification \cite{dai2021coatnet, zhai2021scaling}, text summarization \cite{aghajanyan2020better}, and name entity recognition \cite{yu2020named, yamada2020luke}.

In recent times, deep learning models have become the ``go-to'' methods for supervised and unsupervised tasks involving numerical data, categorical data, time-series data, and text/image data. It is evident from the survey that there is a surge in the use of deep models after 2016. In this part, we revisit the literature with additional studies that employed different deep learning architectures for analyzing microblog data. The revisit includes only the studies published in 2019 or later. \citeauthor{kang2017prediction}~\cite{kang2017prediction} used a deep neural network for predicting crime occurrences, \citeauthor{han2018nextgen}~\cite{han2018nextgen} proposed a distributed deep anti-money laundering model, and \citeauthor{wei2017convolution}~\cite{wei2017convolution} built a CNN-LSTM model for hate speech detection. \citeauthor{neppalli2018deep}~\cite{neppalli2018deep} developed a CNN model for identifying informative disaster specific tweets, \citeauthor{ray2019keyphrase}~\cite{ray2019keyphrase} presented a biLSTM model for keyphrase extraction, \citeauthor{kumar2019location}~\cite{kumar2019location} used a CNN model for extracting location names from tweets created during emergencies, and \citeauthor{lamsal2021twitter}~\cite{lamsal2021twitter} trained word2vec-based RNN models for disaster response. For analyzing stock markets, \citeauthor{xu2018stock}~\cite{xu2018stock} developed a deep generative model, \citeauthor{khan2020stock}~\cite{khan2020stock} worked on an MLP and \citeauthor{shi2018deepclue}~\cite{shi2018deepclue} proposed a text-based deep model.

Similarly, \citeauthor{zhai2020long}~\cite{zhai2020long} proposed an LSTM model for predicting air quality using Weibo posts, \citeauthor{ali2021traffic}~\cite{ali2021traffic} designed a biLSTM model for traffic accident detection using Twitter and Facebook. \citeauthor{beskow2020evolution}~\cite{beskow2020evolution} designed an LSTM-CNN model for detecting and characterizing political memes, and \citeauthor{ertugrul2019activism}~\cite{ertugrul2019activism} developed an LSTM model for predicting future protests. \citeauthor{serban2019real}~\cite{serban2019real} used a CNN-FastText model for detecting events and tracking disease outbreaks.

Furthermore, neural topic models~\cite{wang2020neural1, wang2020neural2} have been proposed as opposed to mathematically arduous traditional methods such as LDA. Also, there are graph-based neural models~\cite{zhou2020graph} that aim to infer topological relationships from message passing between nodes. These neural models have been explored across various domains, including Few-shot Image Classification, Text Classification, Sequence Labelling, Semantic Extraction, Event Extraction, Fact Verification~\cite{zhou2020graph}. In general, graph representation learning methods have applications across multiple areas: \textit{unsupervised}---graph reconstruction (compress graphs into low-dimensional vectors), link prediction (predicting friendship in social networks), clustering (discovering communities), visualization (qualitative understanding of graph properties); \textit{supervised}---node classification (predict node labels), graph classification (predict graph labels) \cite{chami2020machine}.

\subsection{A Commentative Revisit}
In this section, we carry out a commentative revisit to the six ``thematic" areas.

\textbf{Crimes.} Previous studies have utilized microblog conversations for predicting future crimes, analyzing crime rates, and studying the correlation between masses of people at different venues and the occurrence of real crimes at those venues. Some studies implement the ``grid" strategy for identifying crime hot spots; changing the scale and shape of those grids can produce different results---trade a ``hot spot" as a ``neutral spot"---thus leading to inefficient spatial assignments of resources. Crimes such as murders and vandalisms are organized in nature, and conversations relating to them are not reflected online. Therefore, except for such crimes, researchers have confirmed the contribution of microblog data to a better understanding of future crime scenarios and also have reinforced crime-related microblog contents as ``social crime sensors". However, the literature in this thematic area appears North America-centered; contributions from other parts of the world seem required.

\textbf{Disasters.} Disaster-specific microblog conversations have been used for enhancement of situational awareness and prediction and assessment of disasters such as hurricanes, earthquakes, floods, cyclones, etc. The literature in this area has contributions from researchers globally and has the most comprehensive collections of datasets\footnote{https://crisisnlp.qcri.org/} related to disasters and crisis events. The continuing research in this area has led to the design of the-state-of-the-art social media messages mining system,  AIDR, that analyzes Twitter conversations created during humanitarian crises for disaster relief. The science behind the system has been detailed over a stack of research papers\footnote{https://github.com/qcri-social/AIDR/wiki/The-science-behind-AIDR} published through 2013, depicting the progress in the area over the years. This thematic area, however, has an unexplored avenue of research---the application of data mining on \textit{romanized scripts}.

\textbf{Finance.} The literature in this thematic area confirms the relationship between the conversation generated by financial communities on microblogging platforms with stock pricing, market movements, and sustainable business models. The majority of the studies in this area seem to be focused on time-series analysis, with the Granger-causality test, due to its computational simplicity, continuing as the most favored method for causality analysis. Concerning the stock market analysis, the literature is focused majorly on the US-based stock indexes such as NYSE, NASDAQ, and DJIA. The existing methodologies proposed for these indexes are yet to be verified for their generality to other countries' stock indexes. Also, the current state of research seems to be incorporating only a few variables out of many possible economic variables such as supply and demand, company's state (revenue, debt, an influx of investor capital, etc.), investors' sentiment, central bank's policy on interest rates, politics, current events (protests, civil war, etc.), and foreign exchange. 

\textbf{Physical Environment.} Earlier studies have ascertained microblog users as ``environment monitoring sensors"; their generated online conversations have been observed valuable in producing insights associated with traffic incidents/patterns/conditions, weather situations, air quality, and climate change. Physical monitoring systems cover only certain regions; however, bringing-in online conversations into the picture for extrapolation can yield fine-grained details for areas that are outside the reach of the existing monitoring systems. Next-generation systems should focus on adding new types of cross-domain data relevant to population activities. This research area, especially the traffic-related applications, considering conversations from roads are usually minimal, require designing an efficient geocoder for identifying the near-accurate geographical locations (e.g., road segments) of the origin of non-geotagged conversations.

\textbf{Politics.} Politics-based microblog conversations have been used across studies for studying the popularity of election nominees, predicting election outcomes, discovering political topics, and understanding communication patterns of bots and humans during riots/protests. Such analyses get majorly influenced by rumors, spam, and misinformation. While these kinds of conversations are an issue across all thematic areas, the wave of misinformation content when targeted to the voters can fuel conspiracy theories that could impact the election outcome. Therefore, the early detection of such content, the analysis of the distinctive characteristics of those content, and the study of how they propagate should be the major concerns of future studies.

\textbf{Health and Population.} The literature in this area discusses the use of public-health-related microblog conversations in predicting new epidemic/pandemic cases, extracting detailed insights regarding an illness or abuse, and tracking conspiracy theories and misinformation trails related to COVID-19. Studies performing early forecasts of new cases or the trend rely majorly on the volume of conversations at different levels---tweet count, sentiment-based count, a specific theme-based count. The issue with the ``volume" feature is its reliability; methods based on this feature are highly likely to generate biased forecasts during an avalanche of auto-generated conversations. Granular-level latent topic-based models can address this issue to some extent. The literature suggests that although the conversational models show the applicability of online conversations as an indicator of epidemic/pandemic activity, such models should be considered a supplement but not a substitute for epidemiological models. Researchers globally are applying a wide range of analyses on publicly available COVID-19 discourse. Currently, the primary concerns of researchers in the Crisis Computing domain is to make early forecasts of possible cases while incorporating social media variables into the forecasting models, identify the needs of a region through discourse mining, and study the propagation of misinformation concerning the disease and its vaccines, such that the proposed methodologies can be generalized to future epidemic/pandemic outbreaks.

Apart from the ones discussed above, there are a few more challenges and avenues for future research that are common across all ``thematic" areas; we briefly outline them below:

\textit{(a) Inadequate context and Data rate limitations.}
Microblog data is easily misinterpreted, especially when context is unavailable. Inferences can be inaccurate due to biases in microblogger demographics towards younger and tech-savoy users, regional/local users being undistinguished from tourists/travellers, private posts being unavailable, and lack of multi-lingual processing. Freely available microblog data is usually rate limited by the service providers, e.g. a few percent of the total available data, or a few million microblogs per day. Broad queries, and techniques like real-time query expansion, can easily saturate these limits leading to <100\% recall. Precise queries, targeting specific information, are therefore important. %Access to complete data can readily cost over US\$100k/year.%

\textit{(b) Ground truth and Multiple data sources.}
Establishing ground truth, to evaluate the effectiveness of a given approach, can be problematic and/or involve an expensive and time-intensive process of labelling data, typically through crowd-sourcing or citizen science. Labelled data may not be available in real-time as a situation unfolds nor in all required languages. \textit{Weak supervision}~\cite{zhou2018brief} is an emerging direction to address this. It is increasingly important to complement microblog data with information received from native sensors and also include data from Google Trends, web-references, Reddit~\cite{manikonda2018twitter}, cellular communications (SMS)~\cite{meier2010unprecedented}, government agencies.

\textit{(c) Knowledge transfer and generalization.}
There is a significant trend towards language-independent nonexclusive models for homogeneous events. E.g. a classification model designed for the 2015 Nepal Earthquake should be effective at analyzing other earthquake events.

\textit{(d) Adversaries and fake information.}
Systems that process public information can be vulnerable to adversaries who attack the system by posting fabricated microblog data specifically to manipulate the outcomes. The increasing deluge of fake information is a serious problem, and detection and resolution of adversaries is an important research direction~\cite{kumar2018false,zubiaga2018detection}. 

\textit{(e) Geo obfuscation/privacy.} 
Precise geo coordinates are extremely valuable for situation awareness, but at the same time microbloggers are placing increased importance on their individual privacy. This leads to omissions and/or obfuscation of geo information; today <1\% of tweets are geotagged~\cite{lamsal2021design}. Use of bounding boxes and toponyms to enhance geo information needs to be balanced with individuals' privacy.
    
\textit{(f) Geo-political issues.}
Many governments are showing an increasing concern over the vulnerability of their citizens and furthermore their national interests, when significant portions of their citizen population are subjected to high-level analysis of their social media activity; particularly when this is undertaken by companies and researchers in foreign countries.

\section{Conclusion}
\label{conclusion}

AI techniques for processing microblog social media data have become a significant research direction with applications over all areas of life. In this survey we have shown the seminal work and state-of-the-art approaches, from research over the last decade, in six thematic areas that cover the majority of compelling applications today: \emph{Crime}, \emph{Disasters}, \emph{Finance}, \emph{Physical Environment}, \emph{Politics}, and \emph{Health and Population}. We provided a novel, integrated methodological perspective of this body of research, in terms of \emph{Perception}, \emph{Comprehension} and \emph{Visualization/Projection}---the three stages of Situation Awareness. Our survey provides an understanding of how researchers collect, pre-process, and analyze microblog data, with emphasis on how text, image, and spatio-temporal data is utilized for modelling and analysis. We observed a surge of interest in deep learning methods for analyzing microblog data and we expect research efforts to continue strongly in this direction. 

Incorporating knowledge from microblog social media data into the three stages of Situation Awareness is a challenging task to automate due a range of issues that we identified in this survey. Challenges stem not only from the intrinsic nature of the data, being largely raw and unstructured, but also from the fact that practical systems, having significant impact and visibility in society, inevitably become the target of cyber-attacks and information manipulation from adversarial third-parties---fake news and misinformation is surging activity that needs to be understood and overcome. As AI systems become more powerful and as social media data becomes more abundant and richer in semantics, we expect an increasing emphasis on the need of AI techniques to suppress the effect of a similarly increasingly intelligent adversarial response. Researchers must concurrently take into account privacy and security of individuals and requirements of governments and law makers, nationally and internationally.

Finally, prior to 2016, the microblog analytics domain was limited in terms of public datasets. However, at present, the number of datasets concerning disasters, protests, politics, propaganda, sports, climate change and health appears to be growing at an unprecedented pace\footnote{https://catalog.docnow.io/} with many areas yet to be addressed. Twitter's new academic endpoint
%\footnote{Twitter's new academic research product track allows approved academic researchers to access its \textit{full-archive endpoint} for collecting tweets from as early as 2006. This has opened up independent research possibilities involving historical events.}%
has also enabled researchers to collect historical tweets at a level surpassing what has been available over the last decade. We hope that more researchers endeavor to curate interesting datasets to support continued research in this area.

\bibliographystyle{ACM-Reference-Format}
\bibliography{ref}

\end{document}